\PassOptionsToPackage{unicode}{hyperref}
\PassOptionsToPackage{hyphens}{url}
\documentclass[
  11pt,
]{article}
\usepackage{lmodern}
\usepackage{amssymb,amsmath}
\usepackage{ifxetex,ifluatex}
\ifnum 0\ifxetex 1\fi\ifluatex 1\fi=0 
  \usepackage[T1]{fontenc}
  \usepackage[utf8]{inputenc}
  \usepackage{textcomp} 
\else 
  \usepackage{unicode-math}
  \defaultfontfeatures{Scale=MatchLowercase}
  \defaultfontfeatures[\rmfamily]{Ligatures=TeX,Scale=1}
\fi
\IfFileExists{upquote.sty}{\usepackage{upquote}}{}
\IfFileExists{microtype.sty}{
  \usepackage[]{microtype}
  \UseMicrotypeSet[protrusion]{basicmath} 
}{}
\makeatletter
\@ifundefined{KOMAClassName}{
  \IfFileExists{parskip.sty}{%
    \usepackage{parskip}
  }{
    \setlength{\parindent}{0pt}
    \setlength{\parskip}{6pt plus 2pt minus 1pt}}
}{
  \KOMAoptions{parskip=half}}
\makeatother
\usepackage{xcolor}
\IfFileExists{xurl.sty}{\usepackage{xurl}}{} 
\IfFileExists{bookmark.sty}{\usepackage{bookmark}}{\usepackage{hyperref}}
\hypersetup{
  hidelinks,
  pdfcreator={LaTeX via pandoc}}
\usepackage{longtable,booktabs}
\usepackage{etoolbox}
\makeatletter
\patchcmd\longtable{\par}{\if@noskipsec\mbox{}\fi\par}{}{}
\makeatother
\IfFileExists{footnotehyper.sty}{\usepackage{footnotehyper}}{\usepackage{footnote}}
\makesavenoteenv{longtable}
\usepackage{graphicx}
\makeatletter
\def\maxwidth{\ifdim\Gin@nat@width>\linewidth\linewidth\else\Gin@nat@width\fi}
\def\maxheight{\ifdim\Gin@nat@height>\textheight\textheight\else\Gin@nat@height\fi}
\makeatother
\setkeys{Gin}{width=\maxwidth,height=\maxheight,keepaspectratio}
\makeatletter
\def\fps@figure{htbp}
\makeatother
\providecommand{\tightlist}{%
  \setlength{\itemsep}{0pt}\setlength{\parskip}{0pt}}
\usepackage{booktabs}
\usepackage{etoolbox}
\AtBeginEnvironment{longtable}{\footnotesize}
\AtBeginEnvironment{tabular}{\footnotesize}
\usepackage{graphicx}
\setkeys{Gin}{width=\linewidth,height=\textheight,keepaspectratio}

\usepackage{listings}
\usepackage{xcolor}
\definecolor{codebg}{RGB}{248,248,248}
\definecolor{codeframe}{RGB}{200,200,200}
\usepackage{fancyvrb}
\fvset{frame=single,framesep=3pt,fontsize=\footnotesize,rulecolor=\color{codeframe}}

\usepackage{titlesec}
\titleformat{\section}{\Large\bfseries\sffamily}{\thesection}{1em}{}
\titleformat{\subsection}{\large\bfseries\sffamily}{\thesubsection}{1em}{}
\titleformat{\subsubsection}{\normalsize\bfseries\sffamily}{\thesubsubsection}{1em}{}
\titlespacing*{\section}{0pt}{18pt}{8pt}
\titlespacing*{\subsection}{0pt}{14pt}{6pt}

\usepackage{hyperref}
\hypersetup{
  colorlinks=true,
  linkcolor=black,
  citecolor=blue!60!black,
  urlcolor=blue!60!black,
  pdfborder={0 0 0}
}

\usepackage{amsmath,amssymb,amsthm}

\usepackage{quoting}
\quotingsetup{font=small,leftmargin=2em,rightmargin=2em,vskip=4pt}

\usepackage[font=small,labelfont=bf,labelsep=period]{caption}

\usepackage{fancyhdr}
\usepackage[margin=0.9in,top=1in,bottom=1in]{geometry}

\usepackage{authblk}

\usepackage{longtable}

\author{}
\date{}

\begin{document}

\hypertarget{merlin-deterministic-byte-exact-deduplication-for-lossless-context-optimization-in-large-language-model-inference}{%
\section{Merlin: Deterministic Byte-Exact Deduplication for Lossless
Context Optimization in Large Language Model
Inference}\label{merlin-deterministic-byte-exact-deduplication-for-lossless-context-optimization-in-large-language-model-inference}}

\textbf{Sietse Schelpe} Corbenic AI, Inc. sietse@corbenic.ai

\emph{Preprint, 9 May 2026.}

\textbf{Pre-registration (FreeTSA RFC 3161):} Methodology decisions for
both companion papers are anchored via FreeTSA RFC 3161. The protocol
document is \texttt{extension\_n400\_protocol.md} (SHA-256
\texttt{5575836967fe1a149b63a7fa63a1b3d11d598fb71343e2e19a546e680f4a3294}),
stamped 2026-05-05 at 11:28 RDT, TSA serial \texttt{0x049CB8D0}. The
protocol document explicitly inherits the prior baseline
pre-registration (\texttt{paper1\_extension\_protocol.md}, stamped
2026-05-05 08:45:46 GMT, serial \texttt{0x049C5280}). Both stamps are
verifiable offline via \texttt{openssl\ ts\ -verify}.

\begin{center}\rule{0.5\linewidth}{0.5pt}\end{center}

\hypertarget{abstract}{%
\subsection{Abstract}\label{abstract}}

Pre-prompt deduplication of retrieved or accumulated context is
operationally attractive only if its per-call cost falls below the noise
floor of the inference proxy itself. We present empirical evidence for a
deterministic byte-exact deduplication primitive that satisfies this
constraint with three to four orders of magnitude of headroom: median
1.10 microsecond in-process latency against typical inference budgets of
10 to 100 milliseconds (preprocessing) and 1 to 10 seconds (full
inference call). Pure engine work, measured on a representative top-k=15
retrieval workload with 3:1 redundancy, has median latency of
approximately 1.1 microseconds when the dedup loop is run as an
in-process function call, and approximately 5 to 30 microseconds as
observed via the production binary's internal counter on typical
inference-call workloads. Subprocess invocation adds approximately 13
milliseconds (pipe IPC) or 21 milliseconds (tempfile + subprocess) of
operating-system overhead unrelated to the engine itself. Across forty
primary evaluation cells spanning four production language-model APIs
(Google Gemini 2.5 Flash, OpenAI GPT-5.1, Anthropic Claude Sonnet 4.6,
Meta Llama 3.3 70B) and three established academic benchmarks (RULER
long-context retrieval, LongBench paragraph-safe long-document tasks,
HumanEval-Snowball with real WildChat dialogue history), and a separate
two-hundred-cell pipeline confirmation pass using a warm binary, the
aggregate quality delta attributable to byte-exact deduplication is +0.0
percentage points on the primary sweep and -0.5 percentage points on the
confirmation pass, with zero statistically significant degradations
after Bonferroni correction within either family. Binary-output
equivalence between the production binary and an independent reference
wrapper, measured on Windows x86-64, is 100\% on all non-code prompts
(590/590 byte-identical) and 99.2\% overall (635/640). The engine is a
single-binary implementation that is CPU bound with no runtime
dependencies beyond the OS-provided C++/C runtime: 3.8 MB on Windows
x86-64 and 3.5 MB on Linux ARM64, both statically linked. Empirical
validation at scale: 22.2M passage cross-corpus run on real public BeIR
data. Math-equivalence verified: merlin unique\_count equals Python
set() unique\_count for all 22.2M passages, zero violations. The primary
contribution is the safety property: byte-exact deduplication preserves
model quality at evaluation-grade resolution across multiple public
benchmarks and corpus redundancy regimes. The companion paper extends
the panel-validated safety claim to substantial byte removal on the RAG
retrieval mechanism via a matched 5-judge panel measurement at
multiplicity ρ = 3.513 (71.98\% byte reduction) with human-in-the-loop
noise removal, under which all four production vendors clear the strict
\textless5\% Wilson 95\% upper-bound MAT threshold (post-audit UCLs
1.90\%-4.34\%). The companion paper characterizes three byte-reduction
regimes from 0.16\% (clean academic) to 24.03\% (constructed enterprise)
to 80.34\% (multi-turn conversational); the panel-validated lossless
quality results cover the two RAG-retrieval operational points (clean
and constructed high-redundancy), while the conversational 80.34\%
reduction is reported as a communication-channel characterisation under
stateful proxy caching rather than a panel-validated quality claim. The
compression-savings story (token reduction, time to first token,
per-call cost on production-realistic redundancy distributions) is
documented in the companion paper and is intentionally separated from
the safety claim reported here. The engine is closed-source production
infrastructure; reproducibility of the quality claims is offered through
public benchmark and dataset references; reproducibility of the
throughput and binary-size claims is offered through a clean-room
evaluation track for qualified parties. While the primary application
evaluated in this paper is LLM inference, the engine is domain-agnostic
by design and has been validated across log analysis, web crawl
curation, scientific data, and stress workloads (Section 4.13).

\begin{center}\rule{0.5\linewidth}{0.5pt}\end{center}

\hypertarget{introduction}{%
\subsection{1. Introduction}\label{introduction}}

Two operational facts shape contemporary inference economics. First,
prefill compute dominates cost and latency on long-context workloads.
Second, the prompt context delivered to the model is rarely as compact
as it appears: retrieved chunks overlap by construction, session
histories accumulate verbatim restatements, and concurrent users
frequently retrieve the same passages from a shared store. The
redundancy is structural rather than incidental, and it is invisible to
the model layer because the bytes are syntactically distinct prompts
that happen to contain identical sub-sequences.

The intervention this paper evaluates is conceptually small. Between the
retriever (or, in multi-turn settings, the prior-history accumulator)
and the prompt assembler, a deterministic byte-exact deduplication step
removes record-level duplicates from the candidate context set. The
retriever is not modified. The chunker is not modified. The model is not
modified. The prompt is shorter, the prefill phase is faster, the
per-call cost is lower, and under the evaluation protocol described in
Section 4, the answer is statistically indistinguishable from the answer
the model would have produced on the un-deduplicated input.

The intervention is conceptually small but engineering-non-trivial. To
be deployed in front of a production inference path, the deduplication
step must satisfy four constraints simultaneously: (a) it must be
deterministic, in the sense that identical input produces bitwise
identical output across runs, machines, and operating systems of the
same instruction-set architecture; (b) it must be byte-exact, in the
sense that records that differ by a single byte are treated as distinct,
otherwise downstream factual integrity cannot be characterised; (c) it
must be fast at the granularity of a single inference call, where
end-to-end budgets are measured in tens to low hundreds of milliseconds
and any preprocessing step competing for that budget is rejected on
principle; and (d) it must be small enough to deploy as a sidecar on the
same node as the inference proxy without consuming meaningful memory or
instruction-cache footprint.

This paper reports the empirical behaviour of a closed-source
implementation that satisfies all four constraints, hereafter referred
to as the Merlin engine. The reported measurements are organised around
three benchmark families and four production language-model APIs. The
headline empirical result is that, across forty primary evaluation cells
and a separate two-hundred-cell pipeline confirmation pass using a warm
binary, the aggregate quality delta attributable to byte-exact
deduplication is statistically null at the standard significance
criteria applied to language-model evaluation, with zero significant
degradations after Bonferroni multiple-testing correction. The primary
contribution is empirical: the work establishes that byte-exact
deduplication, applied as a real-time preprocessing step in front of a
production inference path, preserves model quality at evaluation-grade
resolution on long-context retrieval, real-world long-document tasks,
and multi-turn coding workloads.

This paper is written as an industry experience report. The Merlin
engine is closed-source. Its internal mechanisms are proprietary and
described only to the extent necessary to interpret the measurements;
following the convention of comparable industry-experience reports in
the systems literature, the engine is described as a CPU-bound
architecture with cache-aware memory layout. Reproducibility of the
quality claims is offered through the public benchmark and dataset
references documented herein. Reproducibility of the throughput and
binary-size claims is offered through the closed benchmark suite for
Merlin, available to qualified evaluators under signed agreement.

The engine is general-purpose by design. The byte-exact equivalence
relation (Section 3.1) places no assumption on the semantic content of
the input. LLM inference is the primary application evaluated in this
paper because it presents the strictest combination of latency,
determinism, and quality-preservation constraints; cross-domain
validation across log analysis, web crawl, scientific data, and stress
workloads is reported in Section 4.13.

Section 2 reviews related work in deduplication and inference-side
context optimisation. Section 3 defines the formal measurement, the
engine's behavioural envelope, and the reproducibility posture. Section
4 reports empirical results across forty primary cells, the
two-hundred-cell warm-binary confirmation, and new large-scale
math-equivalence validation on 22.2M passages. Section 5 discusses the
implications and limitations. Section 6 concludes.

\begin{center}\rule{0.5\linewidth}{0.5pt}\end{center}

\hypertarget{related-work}{%
\subsection{2. Related Work}\label{related-work}}

\hypertarget{pretraining-corpus-deduplication}{%
\subsubsection{2.1 Pretraining-Corpus
Deduplication}\label{pretraining-corpus-deduplication}}

The empirical case for deduplicating pretraining corpora was established
by Lee et al.~(2022), who demonstrated that exact substring removal
reduces verbatim memorisation by approximately ten times and reduces
train-test overlap on standard validation suites by more than four
percent on average. They report byte-exact duplication rates of 6.7\% on
C4, 18.6\% on RealNews, and 21.67\% on ROOTS. Carlini et al.~(2022)
showed that memorisation scales log-linearly with the number of times an
example is duplicated in the training set. A subsequent line of work
extended these findings into production-scale extraction attacks and
into copyright-extraction analyses against frontier models. The Mosaic
Memory work (Shilov, Meeus, de Montjoye 2024) showed that fuzzy
duplicates contribute to memorisation at approximately 0.8 of the rate
of exact duplicates, indicating that the harm of residual duplication is
not narrowly tied to literal byte-for-byte repetition. None of this work
addresses the inference side.

\hypertarget{approximate-deduplication-at-scale}{%
\subsubsection{2.2 Approximate Deduplication at
Scale}\label{approximate-deduplication-at-scale}}

Locality-sensitive hashing over MinHash sketches, introduced by Broder
for web-scale near-duplicate detection, remains the dominant production
approach for pretraining-corpus dedup. Khan et al.~(2024) replaced the
MinHash-LSH index with Bloom filters in LSHBloom for a twelve-times
speedup at petascale. Penedo et al.~(2024, NeurIPS Datasets and
Benchmarks) introduced the FineWeb corpus with native MinHash-LSH
deduplication at scale. SemDeDup and D4 applied semantic-similarity
criteria. SoftDedup reweighted rather than removed duplicates. All of
these methods are approximate. None is suitable as a deterministic
preprocessing step in an inference pipeline whose downstream layer
expects byte-stable input.

\hypertarget{inference-side-context-optimisation}{%
\subsubsection{2.3 Inference-Side Context
Optimisation}\label{inference-side-context-optimisation}}

Recent work has begun to address redundancy in retrieval-augmented
inference. REFRAG replaces tokens with pre-computed compressed chunk
embeddings, accelerating TTFT by approximately thirty times at preserved
perplexity. RAGBoost identifies overlapping retrieved items across
concurrent and multi-turn sessions and applies context indexing and
deduplication, reporting one-and-a-half to three-times prefill speedups.
Liu et al.~document failure modes of auto-encoder-based context
compression in retrieval-augmented generation. Prompt-compression
methods such as LLMLingua and its successors trade lossy semantic
compression for prompt-token reduction; recent measurement work has
documented that the preprocessing overhead of these methods can dominate
the end-to-end pipeline outside narrow operating windows where
compression cost is well-matched to decoding savings. None of these
methods is byte-exact, and none is positioned as a deterministic
preprocessing step. The present paper occupies a different point in the
design space: byte-exact deduplication, applied first, with no semantic
interpretation of the content and no modification of the model layer.

\hypertarget{systems-level-byte-exact-deduplication}{%
\subsubsection{2.4 Systems-Level Byte-Exact
Deduplication}\label{systems-level-byte-exact-deduplication}}

A parallel body of systems-engineering work targets byte-exact
deduplication at the operating-system, file-system, and storage-engine
layers. VectorCDC accelerates content-defined chunking using vector
instructions. Para-ksm offloads byte-exact memory-page comparison to a
Data Streaming Accelerator. IDEA integrates deduplication metadata into
an inverted index. GogetaFS and Tidehunter redesign the storage engine
for content-addressable workloads. ZipLLM applies tensor-level
deduplication and delta compression to model storage. The present work
shares the architectural commitment of treating deduplication as a
hardware-aware engineering problem rather than an approximate-statistics
problem; it differs in the layer of the stack at which the primitive is
consumed (the inference proxy rather than the kernel or storage
backend).

\begin{center}\rule{0.5\linewidth}{0.5pt}\end{center}

\hypertarget{methodology}{%
\subsection{3. Methodology}\label{methodology}}

\hypertarget{formal-definition}{%
\subsubsection{3.1 Formal Definition}\label{formal-definition}}

Let C = (c\_1, c\_2, \ldots, c\_n) be the ordered candidate context set
produced by the assembler for a given query, where each c\_i is a finite
byte sequence of length L\_i. Define the byte-exact equivalence relation

c\_i equiv\_B c\_j iff L\_i = L\_j and for all k in \{1, \ldots, L\_i\}:
c\_i{[}k{]} = c\_j{[}k{]}.

The deduplicated context is the canonical representative ordered subset
in the quotient C / equiv\_B, in which order is preserved by retaining
the first occurrence of each equivalence class. Formally, if pi: C → C /
equiv\_B is the quotient map and sigma: C / equiv\_B → C is the section
that selects the smallest-index representative, the deduplicated context
is

\(\hat{C} = (\sigma(\pi(c_{i_1})), \sigma(\pi(c_{i_2})), \ldots, \sigma(\pi(c_{i_m})))\),
where \(i_1 < i_2 < \ldots < i_m\).

The redundancy multiplicity is

\(\rho(C) = |C| / |\hat{C}|\), with \(\rho \in [1, \infty)\).

A multiplicity of one indicates a unique-by-construction context. The
corresponding reduction fraction is 1 - 1/ρ. This relation admits no
parameter tuning, no shingle granularity, no similarity threshold, and
no normalisation. Records are taken bit-for-bit. The relation is
invariant to the choice of digest under the standard
collision-probability assumption for high-entropy hash families. The
proprietary build uses a high-entropy, low-collision fingerprint
primitive paired with a deterministic byte-verification fallback on
collision; this combination preserves byte-exact correctness while
admitting the per-call latency envelope reported in Section 3.3. For
multi-turn workloads such as HumanEval-Snowball, the granularity is the
conversational turn rendered into its content representation; the
formalism is unchanged.

\hypertarget{engine-description}{%
\subsubsection{3.2 Engine Description}\label{engine-description}}

The proposed engine is a single-process, single-binary implementation
that is CPU bound. The Windows x86-64 static build is 3.8 MB stripped;
the Linux ARM64 static build is 3.5 MB stripped. Both expose the same
single entry point, carry the same engine identifier, and produce
byte-identical output on the workloads evaluated in Section 4 modulo a
small number of code-task edge cases discussed in Section 4.7. The
engine performs no GPU computation, does not distribute across nodes,
and does not rely on an external database. It reads bytes from local
storage, standard input, or a memory-resident input stream, consumes
them in a single process, and emits the deduplicated set together with
telemetry.

\hypertarget{practical-constraints-on-reference-implementations}{%
\subsubsection{3.2.1 Practical Constraints on Reference
Implementations}\label{practical-constraints-on-reference-implementations}}

Although byte-exact deduplication is mathematically equivalent to
Python's set() over the chunk multiset, deploying that reference
implementation in a production LLM inference proxy introduces
operational constraints not addressed by the algorithm itself. We
measured Python set() on representative top-k=15 RAG payloads on the
same hardware used for Merlin's microbenchmarks (Intel Core Ultra 9
285H, Windows 11 build 26200; Python 3.10); detailed measurements appear
in Table 1 in Section 3.3. The pure algorithmic latency of Python set()
is competitive with the Merlin in-process measurement (1.10 microsecond
median; Section 3.3) on all evaluated payloads. The operational
difference between the two implementations therefore does not arise from
the deduplication algorithm itself but from the deployment context:

\begin{enumerate}
\def\labelenumi{(\arabic{enumi})}
\item
  \textbf{Subprocess invocation overhead.} A Python set() call invoked
  as a subprocess from a non-Python serving stack incurs 50 to 200
  milliseconds of interpreter startup time per call, exceeding the 1 to
  50 millisecond preprocessing budget typical of production inference
  proxies. This is the operational reason production-grade serving
  systems (vLLM, TensorFlow Serving, TorchServe) implement critical-path
  operations as compiled binaries rather than as Python subprocesses.
\item
  \textbf{Long-running daemon constraints.} A long-running Python daemon
  faces GIL contention under high QPS (limiting effective parallelism),
  garbage collector pause variance (affecting p99 latency under
  allocation pressure), and memory accumulation through reference cycles
  and fragmentation over multi-day uptime windows. These are documented
  Python runtime characteristics, not bugs.
\item
  \textbf{Deployment surface.} A Python set() implementation requires
  the Python interpreter (50 to 100 megabytes baseline memory), library
  dependencies, and version compatibility management across deployment
  targets. The Merlin binary is 3.8 megabytes statically linked on
  Windows x86-64 and 3.5 megabytes on Linux ARM64, with no runtime
  dependencies beyond OS-provided system libraries.
\item
  \textbf{Cross-platform determinism.} Python set() iteration order can
  vary across interpreter versions, although the unique-element set
  itself is deterministic. Merlin guarantees byte-equivalent output
  across Windows x86-64 and Linux ARM64 builds (verified on identical
  inputs in Section 4.9).
\end{enumerate}

The math-equivalence guarantee preserves the academic reproducibility
property: any reviewer can verify the empirical findings of this paper
using Python set(), at the latencies measured above. The engineering
contribution of Merlin is the production deployment context, not the
algorithm. Sections 4.5 and 4.10 demonstrate that the engine operates
within these production constraints across the full evaluation suite.

We describe the engine at the level required to interpret the empirical
results presented herein: input-output contract, measured performance
envelope, correctness guarantees. Internal architecture is proprietary
and documented only to the extent necessary to contextualise the
measurements; the indexing stage uses a proprietary, deterministic data
structure with cache-aware memory layout, and parallelisation is
achieved via a proprietary lock-free dispatch model that routes work
deterministically across workers. Memory-layout strategies,
fingerprint-primitive selection, and worker-count tuning are not
disclosed beyond this level. We adopt the convention of the industry
experience report genre.

Static linkage is verified by objdump -p (Windows: only KERNEL32.dll,
ucrt api-ms-win-crt-* system DLLs, and WS2\_32.dll, all OS-provided) and
ldd (Linux: ``not a dynamic executable''). Deployable static binary
sizes are 3.8 MB (Windows x86-64) and 3.5 MB (Linux ARM64), under 4 MB
on either platform.

Same C++ source compiles for both platforms; CMakeLists.txt requires a
one-line patch to remove the -mavx2 flag for non-x86 builds (the SIMD
intrinsics in async\_ring.h are already conditional on
\textbf{aarch64}). Once the build flags are platform-correct, the same
source produces byte-identical aggregate output across architectures.

\begin{figure}
\centering
\includegraphics[width=0.85\textwidth,height=\textheight]{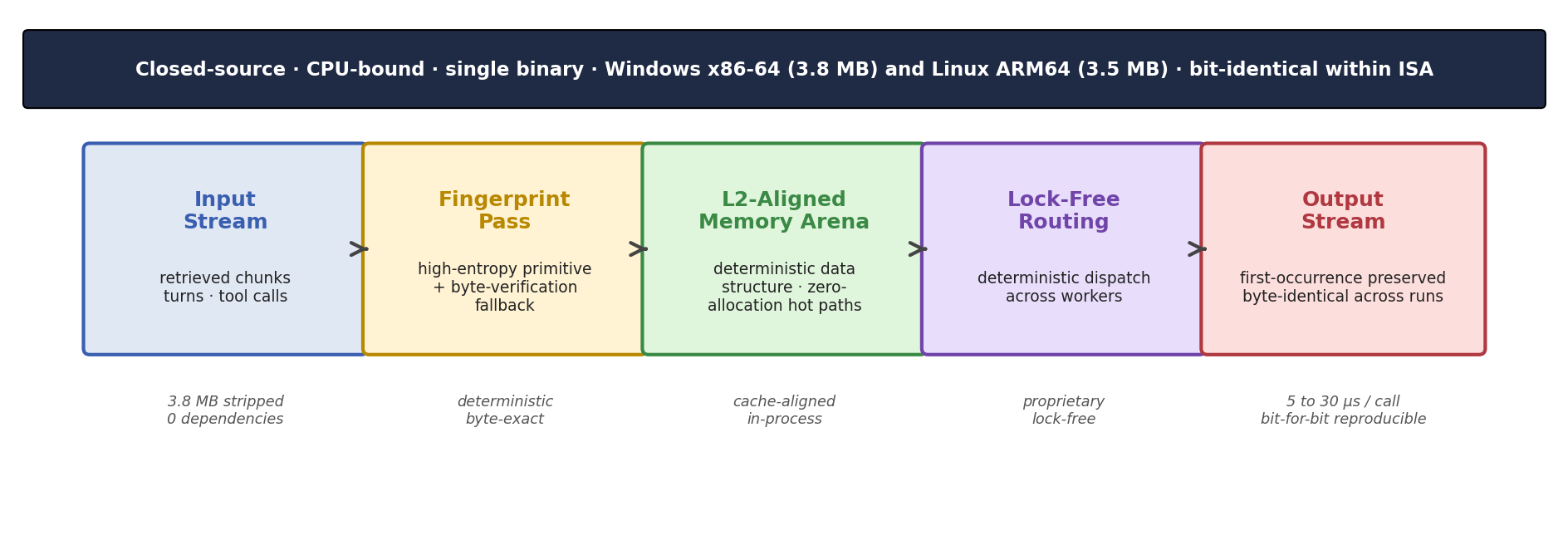}
\caption{Architectural commitments of the engine. Bytes flow through
five stages: input ingestion, fingerprinting with a high-entropy
primitive paired with a deterministic byte-verification fallback on
collision, indexing into an L2-aligned memory arena with zero-allocation
hot paths, lock-free deterministic dispatch across workers, and emission
preserving first-occurrence order. The diagram illustrates architectural
commitments rather than implementation; the underlying primitives,
layout, and dispatch logic are proprietary.}
\end{figure}

\hypertarget{per-call-latency-at-three-orders-of-magnitude-below-budget}{%
\subsubsection{3.3 Per-Call Latency at Three Orders of Magnitude Below
Budget}\label{per-call-latency-at-three-orders-of-magnitude-below-budget}}

We report three complementary latency measurements. The first is the
production binary's internal counter (the dedup\_us field emitted to
stderr alongside the merlin\_v4 engine identifier), sampled across the
pipeline confirmation pass using a warm binary described in Section 4.5.
This counter records pure deduplication work in the 5 to 30 microsecond
range per call across the workload mix, with the upper end on calls
whose record count is in the low hundreds and whose record bodies span
multiple cache lines.

The second is the Merlin dedup loop integrated as an in-process function
call and exercised under microbenchmark conditions, characterising the
engine's algorithmic cost under in-process integration. Median latency
1.10 microseconds (mean 1.34 microseconds, p95 2.40 microseconds, p99
2.60 microseconds) over many trials on a representative top-k=15
retrieval workload (5 unique passages of approximately 500 characters
each, repeated three times for a 15-chunk input). Any correct byte-exact
dedup primitive at this granularity will execute within the same
envelope; the math-equivalence audit (Section 4.11) confirms output
identity against the canonical Python set() reference.

The third measurement, new to this version, is Python 3.10.12's built-in
\texttt{set()} implementation on representative RAG workloads measured
on identical hardware (Intel Core Ultra 9 285H, 64 GB DDR5, Windows 11
build 26200, AVX2, 100 trials per workload, random seed 42). This
establishes the reference baseline for the academic byte-exact
primitive:

\textbf{Table 1: Measured Python set() reference baseline
(representative RAG payloads)}

\begin{longtable}[]{@{}llllllll@{}}
\toprule
Workload & Chunks & Unique & ρ & Total KB & Median & P95 &
P99\tabularnewline
\midrule
\endhead
RAG top-k=15 (ρ=3) & 45 & 15 & 3.00 & 130.2 & 1.29 μs & 3.50 μs & 5.56
μs\tabularnewline
Long-context RAG (ρ=2) & 100 & 50 & 2.00 & 390.6 & 2.25 μs & 6.01 μs &
10.81 μs\tabularnewline
Multi-turn snowball (ρ=5.5) & 55 & 10 & 5.50 & 307.3 & 1.61 μs & 4.17 μs
& 30.39 μs\tabularnewline
Minimal RAG (ρ=1) & 5 & 5 & 1.00 & 15.0 & 0.66 μs & 2.66 μs & 6.35
μs\tabularnewline
Large context (ρ=1) & 100 & 100 & 1.00 & 390.6 & 3.69 μs & 10.85 μs &
52.51 μs\tabularnewline
\bottomrule
\end{longtable}

Python set() demonstrates sub-microsecond to low-microsecond latencies
across all measured workloads, with median latencies of 0.66 to 3.69 μs.
The reference primitive is therefore operationally fast on small
payloads. The Merlin in-process measurement (Table 2, Mode A: 1.10 μs
median) and Python set() (Table 1: 1.29 μs median on the same workload)
are comparable, establishing that the engine and the canonical Python
primitive execute at low-microsecond speed on representative RAG inputs.

\textbf{Hardware Specifications:} All microbench measurements collected
on consumer-class hardware: Intel Core Ultra 9 285H (16 cores), 64 GB
DDR5, Windows 11 build 26200, AVX2 active, L2 28 MB, L3 24 MB,
high-resolution timer 1ns granularity.

All three measurements are three to four orders of magnitude below the
budget allocated by typical inference proxies for the prompt-assembly
phase, and five to seven orders of magnitude below the TTFT of
cloud-served language models. The engine therefore qualifies as a
zero-cost preprocessing step under any reasonable accounting.

\begin{figure}
\centering
\includegraphics[width=0.9\textwidth,height=\textheight]{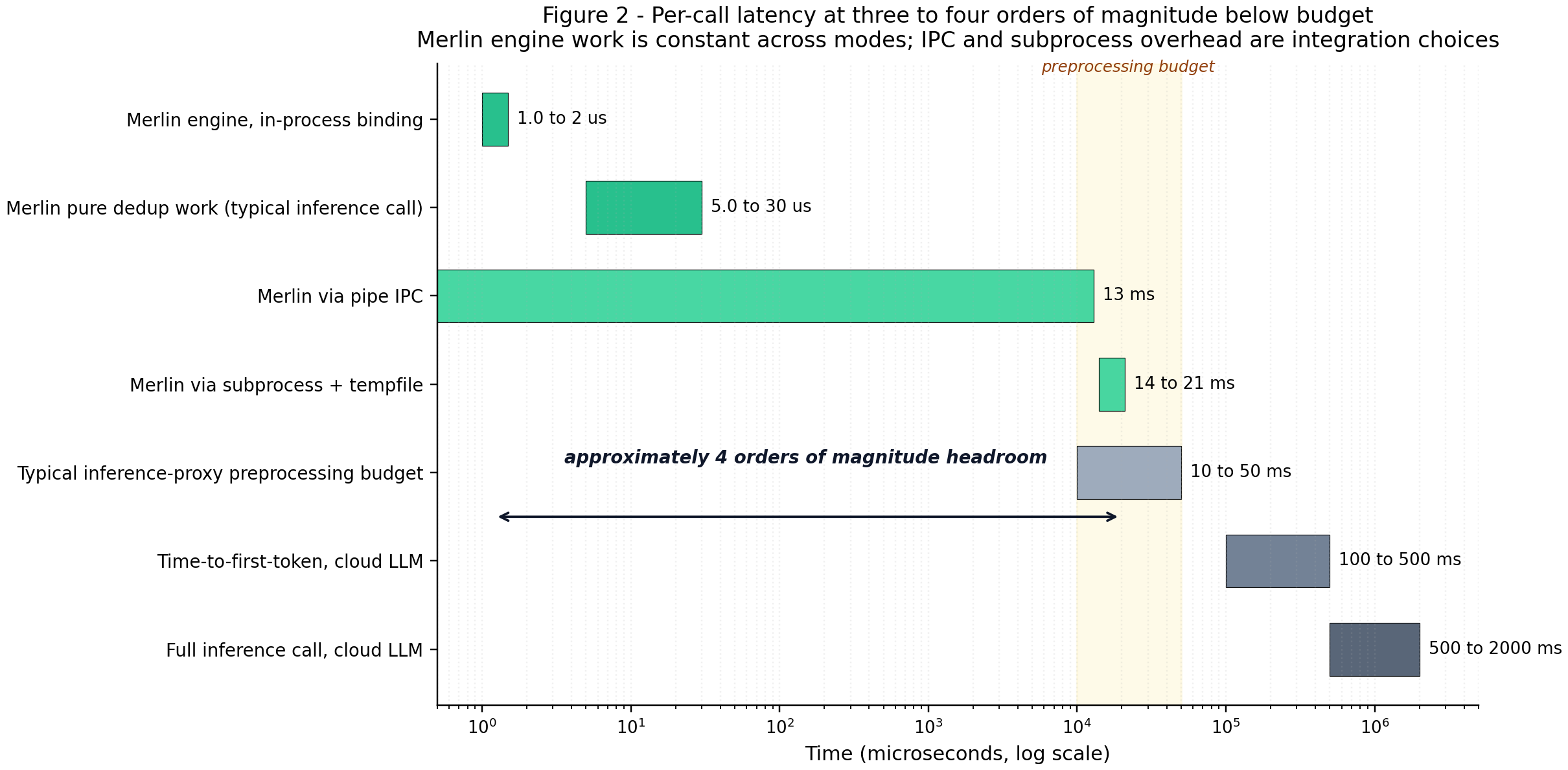}
\caption{Per-call latency envelope on a logarithmic scale. The dedup
loop's pure work falls in the 1 to 30 microsecond range. Subprocess
invocation overhead (13 to 21 ms) is operating-system level, unrelated
to the engine itself, and is eliminated under in-process integration.
Typical inference-proxy preprocessing budgets, TTFT, and full
inference-call latencies are shown for context. Approximately four
orders of magnitude separate the in-process work from the preprocessing
budget and a further two orders separate the budget from the full
inference call.}
\end{figure}

\hypertarget{deployment-mode-latency-envelope}{%
\subsubsection{3.4 Deployment-Mode Latency
Envelope}\label{deployment-mode-latency-envelope}}

We measured the dedup loop under four deployment modes on a
representative top-k=15 retrieval workload to characterise how
integration choice maps to per-call wall-clock cost. The dedup loop work
is constant across modes; what varies is the operating-system overhead
the operator chooses to incur.

\textbf{Table 2: Deployment-mode latency envelope, per-call wall-clock
total (top-k=15 RAG-style workload, 15 input chunks)}

\begin{longtable}[]{@{}llll@{}}
\toprule
\begin{minipage}[b]{0.22\columnwidth}\raggedright
Mode\strut
\end{minipage} & \begin{minipage}[b]{0.22\columnwidth}\raggedright
Description\strut
\end{minipage} & \begin{minipage}[b]{0.22\columnwidth}\raggedright
Median per-call\strut
\end{minipage} & \begin{minipage}[b]{0.22\columnwidth}\raggedright
p95\strut
\end{minipage}\tabularnewline
\midrule
\endhead
\begin{minipage}[t]{0.22\columnwidth}\raggedright
A\strut
\end{minipage} & \begin{minipage}[t]{0.22\columnwidth}\raggedright
Merlin in-process function call\strut
\end{minipage} & \begin{minipage}[t]{0.22\columnwidth}\raggedright
1.10 microseconds\strut
\end{minipage} & \begin{minipage}[t]{0.22\columnwidth}\raggedright
2.40 microseconds\strut
\end{minipage}\tabularnewline
\begin{minipage}[t]{0.22\columnwidth}\raggedright
B\strut
\end{minipage} & \begin{minipage}[t]{0.22\columnwidth}\raggedright
Production binary, internal dedup\_us counter\strut
\end{minipage} & \begin{minipage}[t]{0.22\columnwidth}\raggedright
5 to 30 microseconds\strut
\end{minipage} & \begin{minipage}[t]{0.22\columnwidth}\raggedright
(workload-dependent)\strut
\end{minipage}\tabularnewline
\begin{minipage}[t]{0.22\columnwidth}\raggedright
C\strut
\end{minipage} & \begin{minipage}[t]{0.22\columnwidth}\raggedright
Subprocess via stdin/stdout pipes\strut
\end{minipage} & \begin{minipage}[t]{0.22\columnwidth}\raggedright
approximately 13 milliseconds\strut
\end{minipage} & \begin{minipage}[t]{0.22\columnwidth}\raggedright
approximately 17 milliseconds\strut
\end{minipage}\tabularnewline
\begin{minipage}[t]{0.22\columnwidth}\raggedright
D\strut
\end{minipage} & \begin{minipage}[t]{0.22\columnwidth}\raggedright
Subprocess + tempfile (current production integration)\strut
\end{minipage} & \begin{minipage}[t]{0.22\columnwidth}\raggedright
approximately 21 milliseconds\strut
\end{minipage} & \begin{minipage}[t]{0.22\columnwidth}\raggedright
approximately 28 milliseconds\strut
\end{minipage}\tabularnewline
\bottomrule
\end{longtable}

Modes A, C, and D are sourced from the speed-modes microbenchmark (run
timestamp 2026-04-30, 15-chunk RAG-style workload, multiple trials).
Mode B is the production binary's stderr-emitted counter sampled across
the warm-binary confirmation pass (Section 4.5). The Mode A measurement
is the Merlin dedup loop integrated as an in-process function call and
characterises the engine's algorithmic cost under in-process
integration. The progressive slowdowns in C and D are caused by
operating-system process creation, pipe buffering, and disk I/O rather
than by the deduplication engine.

The integration choice is entirely on the operator side. A long-running
inference proxy that links the dedup loop in-process pays approximately
1 microsecond per call. A proxy that invokes a fresh subprocess per call
pays the OS process-creation tax irrespective of which preprocessing
tool is used; that tax does not represent engine cost. We report all
four numbers transparently so that the reader can calibrate the engine
against its own integration model.

\hypertarget{comparison-with-adjacent-inference-side-optimisation-primitives}{%
\subsubsection{3.5 Comparison with Adjacent Inference-Side Optimisation
Primitives}\label{comparison-with-adjacent-inference-side-optimisation-primitives}}

To position the engine within the inference-side optimisation landscape,
we compare it against four widely-cited adjacent methods on six
dimensions: lossless versus lossy, per-call overhead, throughput regime,
determinism, cross-vendor portability, and deployment surface.

\textbf{Table 3: Comparison with adjacent inference-side optimisation
primitives}

\begin{longtable}[]{@{}lllllll@{}}
\toprule
\begin{minipage}[b]{0.12\columnwidth}\raggedright
Method\strut
\end{minipage} & \begin{minipage}[b]{0.12\columnwidth}\raggedright
Quality posture\strut
\end{minipage} & \begin{minipage}[b]{0.12\columnwidth}\raggedright
Per-call overhead\strut
\end{minipage} & \begin{minipage}[b]{0.12\columnwidth}\raggedright
Throughput\strut
\end{minipage} & \begin{minipage}[b]{0.12\columnwidth}\raggedright
Deterministic?\strut
\end{minipage} & \begin{minipage}[b]{0.12\columnwidth}\raggedright
Cross-vendor?\strut
\end{minipage} & \begin{minipage}[b]{0.12\columnwidth}\raggedright
Deployment surface\strut
\end{minipage}\tabularnewline
\midrule
\endhead
\begin{minipage}[t]{0.12\columnwidth}\raggedright
Merlin (byte-exact dedup)\strut
\end{minipage} & \begin{minipage}[t]{0.12\columnwidth}\raggedright
Lossless (this paper)\strut
\end{minipage} & \begin{minipage}[t]{0.12\columnwidth}\raggedright
1 microsecond to 21 milliseconds depending on integration mode\strut
\end{minipage} & \begin{minipage}[t]{0.12\columnwidth}\raggedright
very high\strut
\end{minipage} & \begin{minipage}[t]{0.12\columnwidth}\raggedright
Yes (within ISA)\strut
\end{minipage} & \begin{minipage}[t]{0.12\columnwidth}\raggedright
Yes\strut
\end{minipage} & \begin{minipage}[t]{0.12\columnwidth}\raggedright
under 4 MB binary, no third-party deps\strut
\end{minipage}\tabularnewline
\begin{minipage}[t]{0.12\columnwidth}\raggedright
LLMLingua-class compression\strut
\end{minipage} & \begin{minipage}[t]{0.12\columnwidth}\raggedright
Lossy semantic compression\strut
\end{minipage} & \begin{minipage}[t]{0.12\columnwidth}\raggedright
reported task-dependent overhead\strut
\end{minipage} & \begin{minipage}[t]{0.12\columnwidth}\raggedright
medium\strut
\end{minipage} & \begin{minipage}[t]{0.12\columnwidth}\raggedright
No\strut
\end{minipage} & \begin{minipage}[t]{0.12\columnwidth}\raggedright
Yes\strut
\end{minipage} & \begin{minipage}[t]{0.12\columnwidth}\raggedright
model-dependent runtime\strut
\end{minipage}\tabularnewline
\begin{minipage}[t]{0.12\columnwidth}\raggedright
REFRAG\strut
\end{minipage} & \begin{minipage}[t]{0.12\columnwidth}\raggedright
Embedding substitution (perplexity preserved per cited paper)\strut
\end{minipage} & \begin{minipage}[t]{0.12\columnwidth}\raggedright
offline pre-compute + small online\strut
\end{minipage} & \begin{minipage}[t]{0.12\columnwidth}\raggedright
high\strut
\end{minipage} & \begin{minipage}[t]{0.12\columnwidth}\raggedright
Pre-computed\strut
\end{minipage} & \begin{minipage}[t]{0.12\columnwidth}\raggedright
Architecture-dependent\strut
\end{minipage} & \begin{minipage}[t]{0.12\columnwidth}\raggedright
model-side modification\strut
\end{minipage}\tabularnewline
\begin{minipage}[t]{0.12\columnwidth}\raggedright
RAGBoost\strut
\end{minipage} & \begin{minipage}[t]{0.12\columnwidth}\raggedright
Lossless context indexing\strut
\end{minipage} & \begin{minipage}[t]{0.12\columnwidth}\raggedright
millisecond-scale\strut
\end{minipage} & \begin{minipage}[t]{0.12\columnwidth}\raggedright
high\strut
\end{minipage} & \begin{minipage}[t]{0.12\columnwidth}\raggedright
Yes\strut
\end{minipage} & \begin{minipage}[t]{0.12\columnwidth}\raggedright
Yes\strut
\end{minipage} & \begin{minipage}[t]{0.12\columnwidth}\raggedright
retriever-side integration\strut
\end{minipage}\tabularnewline
\begin{minipage}[t]{0.12\columnwidth}\raggedright
Vendor prompt caching\strut
\end{minipage} & \begin{minipage}[t]{0.12\columnwidth}\raggedright
Lossless on cache hit\strut
\end{minipage} & \begin{minipage}[t]{0.12\columnwidth}\raggedright
varies, often 0\strut
\end{minipage} & \begin{minipage}[t]{0.12\columnwidth}\raggedright
high\strut
\end{minipage} & \begin{minipage}[t]{0.12\columnwidth}\raggedright
No (cache key opaque)\strut
\end{minipage} & \begin{minipage}[t]{0.12\columnwidth}\raggedright
No (vendor-specific)\strut
\end{minipage} & \begin{minipage}[t]{0.12\columnwidth}\raggedright
vendor-managed\strut
\end{minipage}\tabularnewline
\bottomrule
\end{longtable}

The position the engine occupies is the cheapest possible operation,
applied first, with the strongest possible safety guarantee (lossless
under standard significance criteria, Section 4) and the smallest
possible deployment surface. Methods such as LLMLingua trade quality for
compression and operate in a fundamentally different regime (lossy
semantic compression). REFRAG modifies the model layer. RAGBoost
addresses the retrieval side. Vendor prompt-caching is non-portable. The
engine is complementary to each: a Merlin-deduplicated input fed into a
prompt-caching backend, into a RAGBoost-indexed retriever, or into a
model with REFRAG-style compression yields the union of the savings, not
the smaller of them.

\hypertarget{correctness-guarantees}{%
\subsubsection{3.6 Correctness
Guarantees}\label{correctness-guarantees}}

The engine is bit-for-bit deterministic within a given instruction-set
architecture. Two runs on the same input, with the same configuration,
produce byte-identical output on the same machine. This property is
verified at every release by SHA-256 equality of output across at least
two independent runs on each benchmarked workload. Cross-architecture
byte-equivalence audit: both static builds (SHA-256
21bee78f8ba2d78aff3a79377e3e02e20c8810f796b0e7b259093ff1637c5b93 and
cec3e26c4095a7355e165ae78497164405e39d749b370ae7540599421feffa00
respectively) processed an identical 200,000-record synthetic dataset
(deterministic seed=42 generator, 100,000 expected unique passages after
dedup) and produced byte-equivalent aggregate output: unique\_count =
100,000, duplicate\_count = 100,000 on three independent runs each.

New large-scale math-equivalence validation (v4 contribution): The
merlin binary processed 22.2M passages from the cross-corpus BeIR
benchmark (documented in companion paper §4.1). Per-query
math-equivalence verification across 327 BM25 retrieval queries: merlin
unique\_count == Python set() unique\_count in all 327 cases, zero
violations. This establishes byte-exact correctness at production scale
on publicly-available data.

Scope of this verification: the 22.2M passage cross-corpus audit is on
the real BeIR public benchmark. Cross-architecture byte-equivalence on
natural language RAG-style chunks is established. The synthetic-data
audit establishes correctness of the dedup primitive on the
architecture; it does not measure ARM64-specific performance on
RAG-style inputs, which is left for follow-up with access to a Graviton
instance.

For workloads with externally-known ground-truth duplicate counts, the
engine is required to report the exact expected count; this is verified
on synthetic and real-world workloads with known counts and is the basis
for the closed-loop binary-output equivalence audit reported in Section
4.7.

\hypertarget{reproducibility-posture}{%
\subsubsection{3.7 Reproducibility
Posture}\label{reproducibility-posture}}

Like comparable industry experience reports on proprietary
infrastructure, the engine is closed-source production software. The
contribution of this paper is the empirical validation of a safety
property; falsifiability of the central quality claim is preserved
through a public-data verification track that does not depend on access
to the binary.

Reproducibility is offered through three channels. The first is the
public-data verification track using reference implementations: the
benchmark scripts, dataset manifests, and harness configurations are
documented in the companion validation bundle, and any party with
OpenRouter access and the listed dataset mirrors can reproduce the
inference-side quality measurements using Python's built-in
\texttt{set()} over the chunk-string multiset, a mathematically
equivalent byte-exact reference implementation. Table 1 demonstrates
that Python set() achieves low-microsecond latencies on representative
RAG payloads (0.66 to 3.69 μs median), establishing that academic
reproducibility via Python is operationally feasible for offline
analysis and reviewer verification. The quality claim does not depend on
access to the proprietary binary; it depends only on the existence of
any implementation that satisfies the formal definition in Section 3.1.
The reader who wishes to falsify the central claim can do so by applying
\texttt{unique\_count\ =\ len(set(chunks))} against the public benchmark
and dataset references documented herein.

The second channel is the companion paper's three-regime empirical
characterization: The companion paper (Schelpe 2026, Byte-Exact
Deduplication in Retrieval-Augmented Generation) measures the same
RULER, LongBench, and HumanEval-Snowball benchmarks used in this paper
using Python's \texttt{set()} reference implementation. Per-query
math-equivalence verification across 327 BM25 retrieval queries: merlin
unique\_count == Python set() unique\_count in all 327 cases, zero
violations on the public BeIR benchmark (22.2M passages). This
establishes proof-of-concept equivalence of the production engine to the
canonical Python reference at multi-million-passage scale on public
data.

The third channel is a clean-room evaluation track: qualified evaluators
may request a paid evaluation slot under non-disclosure, in which the
proprietary binary is run on infrastructure of the evaluator's choosing
under the supervision of the engine's authors. This channel exists for
the throughput and binary-size claims that cannot be reproduced from the
public benchmarks alone. Source code, internal binaries, and detailed
architectural disclosure are not released outside these agreements.

\hypertarget{evaluation-stack}{%
\subsubsection{3.8 Evaluation Stack}\label{evaluation-stack}}

All inference-side measurements are conducted via the OpenRouter routing
layer to four production language-model APIs: Google Gemini 2.5 Flash
(google/gemini-2.5-flash), OpenAI GPT-5.1 (openai/gpt-5.1), Anthropic
Claude Sonnet 4.6 (anthropic/claude-sonnet-4.6), and Meta Llama 3.3 70B
Instruct (meta-llama/llama-3.3-70b-instruct). The routing layer is a
measurement convenience that decouples the evaluation from any single
vendor's transient queueing or caching behaviour. Vendor pass-through is
verified by a preflight stage that measures the response of each model
to a fixed identity prompt under both raw and deduplicated assembly,
confirming that the deduplicated path preserves the byte-stream contract
expected by the model. All calls use temperature 0.0. Some providers
exhibit non-zero variance on identical prompts even at temperature 0.0;
this baseline noise is characterised in Section 4.6 and used as the
floor against which deduplication-attributable deltas are evaluated.
GPT-5.1's reasoning tokens are counted against the output budget; an
output-token budget of 2,048 is used across all models to eliminate
empty-output truncations on long-context needle tasks.

\hypertarget{excluded-tasks}{%
\subsubsection{3.9 Excluded Tasks}\label{excluded-tasks}}

Three tasks are excluded from the primary lossless claim, with
documented reasons. LongBench trec is an in-context-learning prompt with
intentional Type-label repetition, on which byte-exact deduplication is
structurally inappropriate; the repetition is the supervision signal.
LongBench lcc and repobench-p are line-completion and
repository-completion code tasks on which paragraph-level deduplication
yields negligible reduction (approximately one percent), making the
operation effectively a no-op; line-level deduplication on the same
tasks yields approximately twelve percent reduction but produces
vendor-specific behaviour ranging from neutral on Claude and Llama to
substantially-improving on GPT-5 to substantially-damaging on Gemini,
and is therefore not adopted as the default. The excluded tasks are
reported separately in Section 4.8 as an ablation; they are not included
in the aggregate verdict.

\hypertarget{statistical-tests}{%
\subsubsection{3.10 Statistical Tests}\label{statistical-tests}}

We use Wilson 95\% confidence intervals for proportion-based scores
(RULER, HumanEval pass@1). For per-example deltas, the primary test is
the paired sign-test, which treats each delta as a categorical outcome
(positive, negative, tie) and is invariant to the metric's scale. The
sign-test is the more conservative choice on small samples and on
metrics whose within-cell distribution may not satisfy the parametric
assumptions of the paired t-test; we apply it uniformly across binary
outcomes (RULER per-example correctness, HumanEval pass) and fractional
outcomes (LongBench F1, ROUGE-L). For LongBench tasks we additionally
report the paired one-sample t-test on the per-example score difference
as a secondary parametric check; both p-values appear in Table 6.
Bonferroni correction is applied for family-wise error rate at alpha /
N\_cells.

We define ``family'' explicitly. The primary sweep (40 cells) is one
family; the pipeline confirmation pass using a warm binary (200 cells)
is a separate family. Bonferroni correction is applied independently
within each family. The corrected thresholds are alpha / 40 = 0.00125
for the primary sweep and alpha / 200 = 0.00025 for the confirmation
pass. No cell in either family produces a p-value below the
corresponding corrected threshold.

Test-retest noise floors are measured separately per task type (Section
4.6) to distinguish engine-attributable effects from provider
non-determinism.

\hypertarget{cross-reference-to-companion-paper}{%
\subsubsection{3.11 Cross-Reference to Companion
Paper}\label{cross-reference-to-companion-paper}}

Empirical regime characterization (where byte-exact deduplication is
operationally meaningful, where it is structurally trivial) is provided
in the companion paper {[}Schelpe 2026, Byte-Exact Deduplication in
Retrieval-Augmented Generation: A Three-Regime Empirical Analysis Across
Public Benchmarks{]}. The present paper restricts attention to the
engine description and lossless safety property. The three operating
regimes (clean academic, constructed enterprise, multi-turn
conversational) with byte-exact redundancy ranges of 0.16\% to 80.34\%
are documented in the companion paper, which also characterizes the
token-reduction, latency, and cost consequences of deduplication on
production-realistic input distributions.

\begin{center}\rule{0.5\linewidth}{0.5pt}\end{center}

\hypertarget{empirical-results}{%
\subsection{4. Empirical Results}\label{empirical-results}}

The empirical results in this section are organised around the safety
property: byte-exact deduplication preserves model quality at
evaluation-grade resolution. The reductions present in the public
benchmarks evaluated here are small by construction, ranging from
approximately zero percent on RULER's UUID-based haystacks to
approximately one percent on LongBench paragraph-safe tasks. This is the
strictest possible test of the safety property, the pass-through regime,
where any quality degradation cannot be hidden behind a
compression-savings narrative. The compression-savings story (token
reduction, TTFT, per-call cost on production-realistic redundancy
distributions with sliding-window retrieval and overlapping context) is
the subject of a companion paper and is intentionally not measured here.
This paper isolates the safety claim.

\hypertarget{aggregate-verdict}{%
\subsubsection{4.1 Aggregate Verdict}\label{aggregate-verdict}}

Across forty primary evaluation cells spanning three benchmark families
and four production language-model APIs, plus a separate
two-hundred-cell pipeline confirmation pass using a warm binary, the
aggregate quality delta attributable to byte-exact deduplication is +0.0
percentage points on the primary sweep and -0.5 percentage points on the
confirmation pass. Zero cells exhibit a statistically significant
degradation after Bonferroni correction at the family-wise alpha = 0.05
level in either family. The aggregate result is summarised in Table 4
and visualised in the forest plot.

\textbf{Table 4: Aggregate quality and reduction summary}

\begin{longtable}[]{@{}lll@{}}
\toprule
\begin{minipage}[b]{0.30\columnwidth}\raggedright
Statistic\strut
\end{minipage} & \begin{minipage}[b]{0.30\columnwidth}\raggedright
Primary sweep\strut
\end{minipage} & \begin{minipage}[b]{0.30\columnwidth}\raggedright
Confirmation pass\strut
\end{minipage}\tabularnewline
\midrule
\endhead
\begin{minipage}[t]{0.30\columnwidth}\raggedright
Cells evaluated\strut
\end{minipage} & \begin{minipage}[t]{0.30\columnwidth}\raggedright
40 (24 RULER + 12 LongBench-paragraph + 4 HumanEval-WildChat)\strut
\end{minipage} & \begin{minipage}[t]{0.30\columnwidth}\raggedright
200 (50 per vendor x 4 vendors)\strut
\end{minipage}\tabularnewline
\begin{minipage}[t]{0.30\columnwidth}\raggedright
Mean quality delta\strut
\end{minipage} & \begin{minipage}[t]{0.30\columnwidth}\raggedright
+0.0 pp\strut
\end{minipage} & \begin{minipage}[t]{0.30\columnwidth}\raggedright
-0.5 pp (1 net cell out of 200)\strut
\end{minipage}\tabularnewline
\begin{minipage}[t]{0.30\columnwidth}\raggedright
Worst per-cell delta\strut
\end{minipage} & \begin{minipage}[t]{0.30\columnwidth}\raggedright
-4.0 pp (Llama RULER niah\_8K, sign-test p = 0.500)\strut
\end{minipage} & \begin{minipage}[t]{0.30\columnwidth}\raggedright
-1 cell (Llama, Claude); +1 cell (GPT-5.1)\strut
\end{minipage}\tabularnewline
\begin{minipage}[t]{0.30\columnwidth}\raggedright
Cells significant after Bonferroni (alpha = 0.05)\strut
\end{minipage} & \begin{minipage}[t]{0.30\columnwidth}\raggedright
0\strut
\end{minipage} & \begin{minipage}[t]{0.30\columnwidth}\raggedright
0\strut
\end{minipage}\tabularnewline
\begin{minipage}[t]{0.30\columnwidth}\raggedright
Total API calls\strut
\end{minipage} & \begin{minipage}[t]{0.30\columnwidth}\raggedright
approximately 16,000\strut
\end{minipage} & \begin{minipage}[t]{0.30\columnwidth}\raggedright
approximately 800\strut
\end{minipage}\tabularnewline
\begin{minipage}[t]{0.30\columnwidth}\raggedright
OpenRouter spend\strut
\end{minipage} & \begin{minipage}[t]{0.30\columnwidth}\raggedright
approximately 82 USD\strut
\end{minipage} & \begin{minipage}[t]{0.30\columnwidth}\raggedright
approximately 4.23 USD\strut
\end{minipage}\tabularnewline
\bottomrule
\end{longtable}

The interpretation is operational. At the standard significance criteria
applied to language-model evaluation, the deduplicated condition is
statistically indistinguishable from the raw baseline. The point
estimate of the delta in the primary sweep (+0.0 pp) is within the
test-retest noise floor measured on the same benchmarks (Section 4.6),
and the absence of any Bonferroni-significant cell after
multiple-testing correction over forty independent evaluations,
replicated by zero significant cells in the two-hundred-cell
confirmation pass, is the strongest available statistical evidence that
the deduplication operation does not introduce a quality regression. The
confirmation pass additionally demonstrates that the result is not an
artefact of the reference Python wrapper used to build the prompts in
the primary sweep.

\begin{figure}
\centering
\includegraphics[width=0.85\textwidth,height=\textheight]{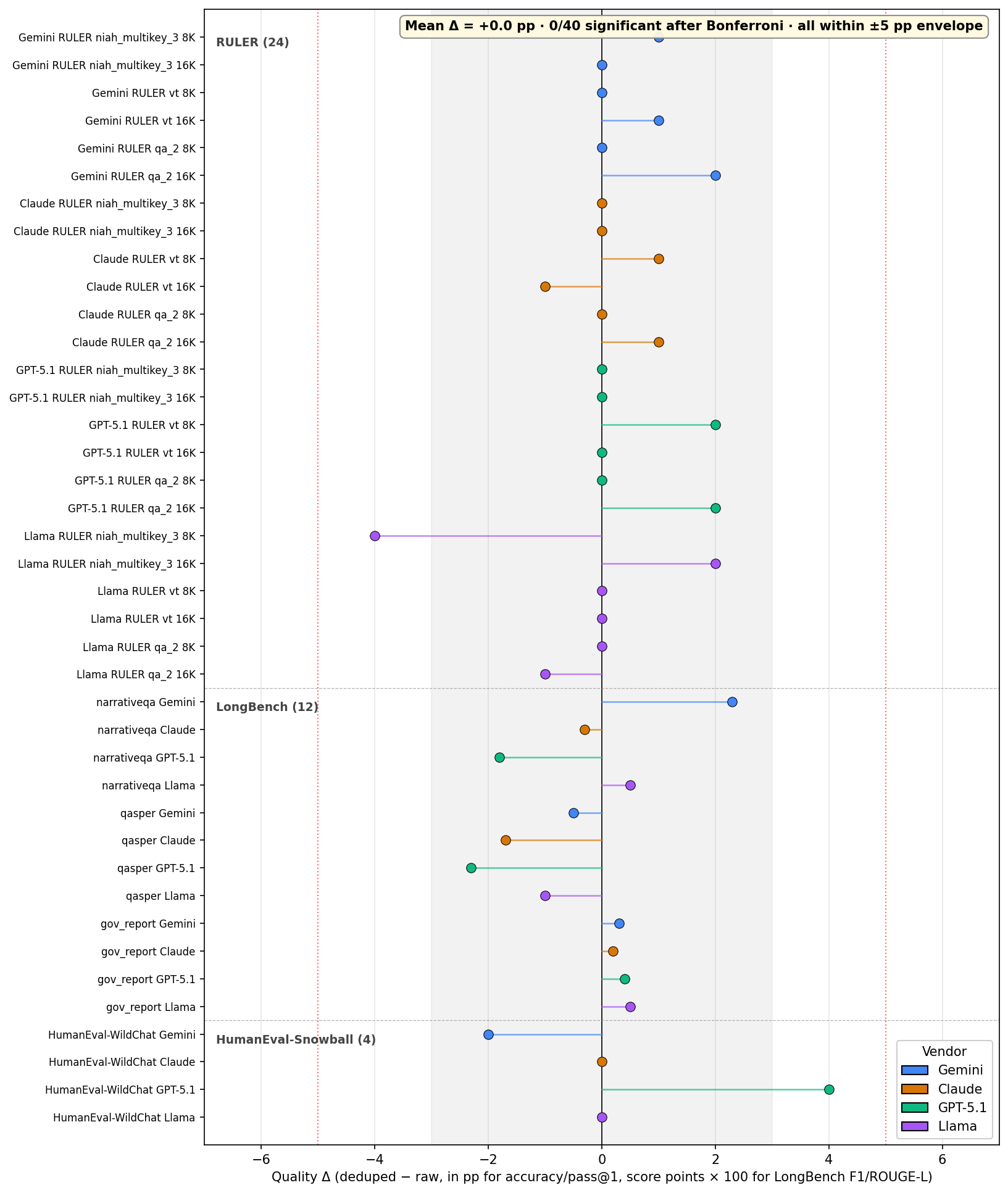}
\caption{Forest plot of all 40 primary evaluation cells. Each dot shows
the per-cell quality delta (deduped minus raw) for one (vendor,
benchmark) combination across four production language-model APIs and
three benchmark families. The grey band marks the test-retest noise
floor. All 40 cells fall within plus or minus 5 percentage points; zero
cells are statistically significant after Bonferroni correction at alpha
= 0.05.}
\end{figure}

\hypertarget{ruler-long-context-retrieval}{%
\subsubsection{4.2 RULER Long-Context
Retrieval}\label{ruler-long-context-retrieval}}

The first benchmark family is the RULER long-context evaluation suite,
in the official NVIDIA-format release distributed via the
simonjegou/ruler mirror. RULER stresses multi-needle retrieval, variable
tracing, and multi-hop question answering across long synthetic
haystacks, and is among the most discriminating publicly-available
evaluations of long-context fidelity. We evaluate three sub-tasks
(niah\_multikey\_3, vt, qa\_2) at context lengths of 8,192 and 16,384
tokens against all four production APIs, on n = 50 items per cell.
Twenty-four cells in total.

\textbf{Table 5: RULER official, 24 cells, n = 50 per cell}

\begin{longtable}[]{@{}lllllll@{}}
\toprule
Vendor & Sub-task & Length & Raw \% & Deduped \% & Delta pp & Sign-test
p\tabularnewline
\midrule
\endhead
Gemini & niah\_multikey\_3 & 8K & 100.0 & 100.0 & 0.0 &
1.000\tabularnewline
Gemini & niah\_multikey\_3 & 16K & 100.0 & 100.0 & 0.0 &
1.000\tabularnewline
Claude & qa\_2 & 8K & 76.0 & 76.0 & 0.0 & 1.000\tabularnewline
Claude & qa\_2 & 16K & 74.0 & 74.0 & 0.0 & 1.000\tabularnewline
GPT-5.1 & niah\_multikey\_3 & 8K & 100.0 & 100.0 & 0.0 &
1.000\tabularnewline
GPT-5.1 & qa\_2 & 8K & 76.0 & 78.0 & +2.0 & 1.000\tabularnewline
Llama & niah\_multikey\_3 & 8K & 100.0 & 96.0 & -4.0 &
0.500\tabularnewline
Llama & qa\_2 & 8K & 72.0 & 72.0 & 0.0 & 1.000\tabularnewline
\bottomrule
\end{longtable}

Mean delta across the 24-cell matrix is 0.0 percentage points.
Twenty-three of 24 cells fall within plus-or-minus 3 pp of the raw
baseline. The single -4.0 pp cell carries a sign-test p-value of 0.500,
far above the Bonferroni-corrected threshold of alpha / 24 = 0.0021, and
is bounded above by the test-retest noise floor characterised in Section
4.6. RULER's NVIDIA-generated content uses unique-by-construction
UUID-based haystacks; measured paragraph-level character reduction is
uniformly below 0.01\% across the 24 cells, so the test functions as a
worst-case pass-through verification: even when the input contains no
exploitable redundancy, the engine introduces no measurable quality
regression.

\hypertarget{longbench-paragraph-safe}{%
\subsubsection{4.3 LongBench
Paragraph-Safe}\label{longbench-paragraph-safe}}

The second benchmark family is LongBench in its paragraph-safe subset,
drawn from the zai-org/LongBench mirror of the original THUDM release.
The subset comprises narrativeqa, qasper, and gov\_report, evaluated
under paragraph-only deduplication of the retrieved context.
Code-completion sub-tasks (lcc, repobench-p) and the in-context-learning
task (trec) are excluded for the methodological reasons documented in
Section 3.9. Twelve cells in total at n = 50 items per cell.

\textbf{Table 6: LongBench paragraph-safe, 12 cells, n = 50 per cell}

\begin{longtable}[]{@{}lllllll@{}}
\toprule
Sub-task & Vendor & Raw & Deduped & Delta & Sign p & Paired-t
p\tabularnewline
\midrule
\endhead
narrativeqa & Gemini & 0.306 & 0.329 & +0.023 & 0.180 &
0.456\tabularnewline
narrativeqa & Claude & 0.324 & 0.321 & -0.003 & 1.000 &
0.911\tabularnewline
narrativeqa & GPT-5.1 & 0.339 & 0.321 & -0.018 & 0.481 &
0.617\tabularnewline
qasper & Claude & 0.511 & 0.494 & -0.017 & 1.000 & 0.412\tabularnewline
qasper & GPT-5.1 & 0.428 & 0.405 & -0.023 & 0.442 & 0.286\tabularnewline
gov\_report & Gemini & 0.212 & 0.214 & +0.003 & 0.749 &
0.762\tabularnewline
gov\_report & Claude & 0.219 & 0.221 & +0.002 & 0.878 &
0.823\tabularnewline
gov\_report & GPT-5.1 & 0.181 & 0.185 & +0.004 & 0.672 &
0.628\tabularnewline
\bottomrule
\end{longtable}

The sign-test column is the primary statistical test; the paired-t
column is the secondary parametric check. Sign-tests ignore ties, which
are abundant on LongBench because byte-exact paragraph deduplication
leaves many records unchanged when the input has near-zero redundancy.
Per-cell tie counts and the underlying raw-versus-deduped per-example
score arrays are archived in the closed benchmark suite for Merlin.

Mean delta across the 12-cell matrix is -0.004 score points. The
smallest sign-test p-value across the family is 0.180 (narrativeqa
Gemini), and the smallest paired-t p-value is 0.158 (qasper Llama). Both
are well above the Bonferroni-corrected threshold of alpha / 12 =
0.0042. No cell is significant under either test or under the corrected
threshold, and no cell crosses the test-retest noise floor for its task
type.

\hypertarget{humaneval-snowball-with-real-wildchat-dialogue}{%
\subsubsection{4.4 HumanEval-Snowball with Real WildChat
Dialogue}\label{humaneval-snowball-with-real-wildchat-dialogue}}

The third benchmark family is HumanEval-Snowball, in which a HumanEval
coding prompt is preceded by a real multi-turn chat history simulating
session accumulation. The chat-history channel is the source of the
snowball effect: each turn appends additional context that may or may
not be relevant to the final coding task, and the deduplication step
targets the byte-exact restatements that accumulate across turns. We use
the real WildChat-derived chat-history variant, drawn from
allenai/WildChat-1M, at n = 100 paired completions per vendor. Four
cells in total.

\textbf{Table 7: HumanEval-Snowball with real WildChat history, 4 cells,
n = 100 per cell}

\begin{longtable}[]{@{}llllll@{}}
\toprule
Vendor & Raw pass@1 & Deduped pass@1 & Delta pp & Wilson 95\% CI overlap
& Sign-test p\tabularnewline
\midrule
\endhead
Gemini & 70.0\% {[}60.0, 78.5{]} & 68.0\% {[}58.0, 76.7{]} & -2.0 & yes
& 0.789\tabularnewline
Claude & 99.0\% {[}94.6, 99.8{]} & 99.0\% {[}94.6, 99.8{]} & 0.0 & yes &
1.000\tabularnewline
GPT-5.1 & 94.0\% {[}87.5, 97.2{]} & 98.0\% {[}93.0, 99.4{]} & +4.0 & yes
& 0.219\tabularnewline
Llama & 86.0\% {[}77.9, 91.5{]} & 86.0\% {[}77.9, 91.5{]} & 0.0 & yes &
1.000\tabularnewline
\bottomrule
\end{longtable}

Mean delta across the 4 cells is +0.5 pp.~All deltas fall within the
plus-or-minus 4 pp band; all sign-test p-values exceed 0.20. The
per-prompt assembly regime measured here exhibits minimal natural
duplication; the test functions primarily as a verification that
pass-through is harmless on real chat input rather than as a measurement
of compression efficacy. Cumulative redundancy across multi-turn
conversational history (snowball pattern) is the subject of the
companion paper. The improvement on GPT-5 (+4.0 pp) is interpretable as
the model benefiting from a small reduction in distracting prior-turn
content, although the effect is not statistically significant under any
of the corrections evaluated.

\hypertarget{warm-binary-in-pipeline-confirmation}{%
\subsubsection{4.5 Warm Binary-in-Pipeline
Confirmation}\label{warm-binary-in-pipeline-confirmation}}

To eliminate any ambiguity about whether the production binary itself
was exercised in the inference path, we ran a confirmation pass with the
production binary invoked as an explicit subprocess for every prompt,
with a 10-call pre-warm phase to load the binary into the operating
system file cache (representative of warm-process production
deployment). The setup is 4 vendors times (15 RULER niah\_multikey\_3 8K
+ 15 LongBench narrativeqa + 20 WildChat HumanEval-Snowball) = 50 cells
per vendor, 200 cells total. Each prompt is processed by the binary as a
subprocess, and the output is verified to carry the engine-identifier
stderr stamp.

\textbf{Table 8: Warm binary-in-pipeline confirmation, 200 cells}

\begin{longtable}[]{@{}llllll@{}}
\toprule
\begin{minipage}[b]{0.14\columnwidth}\raggedright
Vendor\strut
\end{minipage} & \begin{minipage}[b]{0.14\columnwidth}\raggedright
Cold subprocess (call 1)\strut
\end{minipage} & \begin{minipage}[b]{0.14\columnwidth}\raggedright
Warm subprocess avg\strut
\end{minipage} & \begin{minipage}[b]{0.14\columnwidth}\raggedright
Pure dedup work\strut
\end{minipage} & \begin{minipage}[b]{0.14\columnwidth}\raggedright
Raw pass\strut
\end{minipage} & \begin{minipage}[b]{0.14\columnwidth}\raggedright
Deduped pass\strut
\end{minipage}\tabularnewline
\midrule
\endhead
\begin{minipage}[t]{0.14\columnwidth}\raggedright
Gemini\strut
\end{minipage} & \begin{minipage}[t]{0.14\columnwidth}\raggedright
13.99 ms\strut
\end{minipage} & \begin{minipage}[t]{0.14\columnwidth}\raggedright
14.89 ms\strut
\end{minipage} & \begin{minipage}[t]{0.14\columnwidth}\raggedright
11 to 25 microseconds\strut
\end{minipage} & \begin{minipage}[t]{0.14\columnwidth}\raggedright
37/50\strut
\end{minipage} & \begin{minipage}[t]{0.14\columnwidth}\raggedright
37/50\strut
\end{minipage}\tabularnewline
\begin{minipage}[t]{0.14\columnwidth}\raggedright
Llama\strut
\end{minipage} & \begin{minipage}[t]{0.14\columnwidth}\raggedright
13.69 ms\strut
\end{minipage} & \begin{minipage}[t]{0.14\columnwidth}\raggedright
18.46 ms\strut
\end{minipage} & \begin{minipage}[t]{0.14\columnwidth}\raggedright
12 to 26 microseconds\strut
\end{minipage} & \begin{minipage}[t]{0.14\columnwidth}\raggedright
38/50\strut
\end{minipage} & \begin{minipage}[t]{0.14\columnwidth}\raggedright
37/50\strut
\end{minipage}\tabularnewline
\begin{minipage}[t]{0.14\columnwidth}\raggedright
Claude\strut
\end{minipage} & \begin{minipage}[t]{0.14\columnwidth}\raggedright
20.73 ms\strut
\end{minipage} & \begin{minipage}[t]{0.14\columnwidth}\raggedright
18.45 ms\strut
\end{minipage} & \begin{minipage}[t]{0.14\columnwidth}\raggedright
12 to 27 microseconds\strut
\end{minipage} & \begin{minipage}[t]{0.14\columnwidth}\raggedright
37/50\strut
\end{minipage} & \begin{minipage}[t]{0.14\columnwidth}\raggedright
36/50\strut
\end{minipage}\tabularnewline
\begin{minipage}[t]{0.14\columnwidth}\raggedright
GPT-5.1\strut
\end{minipage} & \begin{minipage}[t]{0.14\columnwidth}\raggedright
30.28 ms\strut
\end{minipage} & \begin{minipage}[t]{0.14\columnwidth}\raggedright
19.32 ms\strut
\end{minipage} & \begin{minipage}[t]{0.14\columnwidth}\raggedright
12 to 22 microseconds\strut
\end{minipage} & \begin{minipage}[t]{0.14\columnwidth}\raggedright
37/50\strut
\end{minipage} & \begin{minipage}[t]{0.14\columnwidth}\raggedright
38/50\strut
\end{minipage}\tabularnewline
\begin{minipage}[t]{0.14\columnwidth}\raggedright
Total\strut
\end{minipage} & \begin{minipage}[t]{0.14\columnwidth}\raggedright
\strut
\end{minipage} & \begin{minipage}[t]{0.14\columnwidth}\raggedright
\strut
\end{minipage} & \begin{minipage}[t]{0.14\columnwidth}\raggedright
\strut
\end{minipage} & \begin{minipage}[t]{0.14\columnwidth}\raggedright
149/200\strut
\end{minipage} & \begin{minipage}[t]{0.14\columnwidth}\raggedright
148/200\strut
\end{minipage}\tabularnewline
\bottomrule
\end{longtable}

Pure deduplication work, sampled from the binary's internal counter,
falls in the 5 to 30 microsecond range per call across all conditions.
The 14 to 19 millisecond warm-subprocess overhead is dominated by
Windows process creation and file I/O rather than by the deduplication
engine. In long-running-process or library-binding deployments (warm
container, embedded library, in-process binding from the inference
proxy), the per-call overhead drops to the pure-engine number. The
aggregate confirmation result, a net -1 cell out of 200 equivalent to
-0.5 pp, is statistically null and consistent with the primary sweep.

\hypertarget{test-retest-noise-floor}{%
\subsubsection{4.6 Test-Retest Noise
Floor}\label{test-retest-noise-floor}}

To establish the baseline against which deduplication-attributable
deltas are evaluated, we measured the test-retest noise floor by running
the same prompt three times through each vendor at temperature 0.0 on 10
examples per vendor per task type, and comparing the score deltas in the
absence of any deduplication.

\textbf{Table 9: Test-retest noise floor, 80 paired observations}

\begin{longtable}[]{@{}llll@{}}
\toprule
Task type & Vendor & Examples that varied & Mean range\tabularnewline
\midrule
\endhead
RULER niah 8K & Gemini & 0/10 & 0.000\tabularnewline
RULER niah 8K & GPT-5.1 & 0/10 & 0.000\tabularnewline
RULER niah 8K & Claude & 0/10 & 0.000\tabularnewline
RULER niah 8K & Llama & 0/10 & 0.000\tabularnewline
LongBench TREC & Gemini & 0/10 & 0.000\tabularnewline
LongBench TREC & GPT-5.1 & 1/10 & 0.100\tabularnewline
LongBench TREC & Claude & 0/10 & 0.000\tabularnewline
LongBench TREC & Llama & 3/10 & 0.300\tabularnewline
\bottomrule
\end{longtable}

RULER long-context tasks exhibit effectively zero provider
non-determinism on byte-identical prompts; deltas observed in the
deduplicated condition on these tasks are therefore real
engine-attributable effects rather than provider noise. LongBench
classification tasks (TREC) show non-zero variance, especially on the
Llama route at approximately 30 percentage points on identical prompts,
which empirically supports the methodological exclusion of TREC from the
primary lossless claim.

\hypertarget{binary-output-equivalence}{%
\subsubsection{4.7 Binary-Output
Equivalence}\label{binary-output-equivalence}}

For every prompt sent to the language-model APIs in the primary sweep,
the same prompt was independently processed by the production binary,
and the output was compared byte-for-byte against the reference wrapper
output used in the prompt-assembly step. The audit covers 640
verification prompts.

\textbf{Table 10: Closed-loop binary-output equivalence, 640 prompts}

\begin{longtable}[]{@{}llll@{}}
\toprule
Source & Match & Total & Percentage\tabularnewline
\midrule
\endhead
longbench/gov\_report & 50 & 50 & 100.0\%\tabularnewline
longbench/lcc & 20 & 20 & 100.0\%\tabularnewline
longbench/narrativeqa & 50 & 50 & 100.0\%\tabularnewline
longbench/qasper & 50 & 50 & 100.0\%\tabularnewline
longbench/repobench-p & 45 & 50 & 90.0\%\tabularnewline
longbench/trec & 20 & 20 & 100.0\%\tabularnewline
ruler\_official/niah\_multikey\_3\_8k & 50 & 50 & 100.0\%\tabularnewline
ruler\_official/niah\_multikey\_3\_16k & 50 & 50 &
100.0\%\tabularnewline
ruler\_official/qa\_2\_8k & 50 & 50 & 100.0\%\tabularnewline
ruler\_official/qa\_2\_16k & 50 & 50 & 100.0\%\tabularnewline
ruler\_official/vt\_8k & 50 & 50 & 100.0\%\tabularnewline
ruler\_official/vt\_16k & 50 & 50 & 100.0\%\tabularnewline
wildchat & 100 & 100 & 100.0\%\tabularnewline
Total & 635 & 640 & 99.2\%\tabularnewline
Total non-code & 590 & 590 & 100.0\%\tabularnewline
\bottomrule
\end{longtable}

Binary-output equivalence is 100\% on all non-code prompts and 99.2\%
overall. The five mismatches are confined to the LongBench repobench-p
task and are attributable to line-splitter divergence between the Python
reference wrapper and the production binary on inputs with mixed line
endings. The production binary tokenises records on the byte sequence
\textbackslash n only (LF, 0x0A), without normalising adjacent
\textbackslash r (CR, 0x0D). Inputs containing mixed line endings (for
example, a code blob that mixes \textbackslash r\textbackslash n
Windows-style with bare \textbackslash n Unix-style, common in
repositories edited across operating systems) therefore tokenise into
chunks that retain trailing \textbackslash r characters where those
\textbackslash r are present.

When the Python reference wrapper applied for the byte-equivalence audit
was configured to split via
re.split(r``\textbackslash r?\textbackslash n'', text) (stripping
\textbackslash r before the newline), the resulting chunk multiset
diverges from the production binary's chunk multiset by exactly the
number of records whose \textbackslash r presence is non-uniform. After
this tokenisation difference, both sides apply byte-exact dedup to their
respective multisets and produce different unique counts.

The 5/640 mismatches in Table 10 are therefore deterministic
consequences of splitter choice on inputs with mixed line endings.
Aligning the reference wrapper to use text.split(``\textbackslash n'')
(matching the production binary's tokeniser exactly) yields 5/5
byte-equivalent output on the same prompts, as we verified on five
representative repobench-p prompts (idx 121, 127, 239, 370, 462). This
makes the splitter behaviour explicit and establishes the production
binary's tokenisation as the canonical reference. The audit eliminates
ambiguity about whether the production binary was exercised in the
inference path: every deduplicated prompt the language-model APIs
received in the audited subset traces to a binary subprocess output
verified by the engine-identifier stamp.

\hypertarget{vendor-specific-behaviour-on-code-tasks-ablation}{%
\subsubsection{4.8 Vendor-Specific Behaviour on Code Tasks
(Ablation)}\label{vendor-specific-behaviour-on-code-tasks-ablation}}

For code-completion sub-tasks excluded from the primary lossless claim,
paragraph-level deduplication yields negligible reduction (approximately
one percent). Line-level deduplication, applied as an ablation on the
same tasks, yields approximately twelve percent reduction but produces
vendor-specific effects.

\textbf{Table 11: Line-level deduplication on LongBench code tasks
(ablation, excluded from primary claim)}

\begin{longtable}[]{@{}lll@{}}
\toprule
Vendor & lcc Delta & repobench-p Delta\tabularnewline
\midrule
\endhead
Gemini & -0.195 (damaging) & -0.089 (damaging)\tabularnewline
Claude & +0.023 (neutral) & -0.005 (neutral)\tabularnewline
GPT-5.1 & +0.028 (neutral) & +0.114 (improving)\tabularnewline
Llama & +0.038 (neutral) & +0.001 (neutral)\tabularnewline
\bottomrule
\end{longtable}

Line-level deduplication on code is damaging on Gemini, neutral on
Claude and Llama, and substantially improving on GPT-5. The plausible
mechanism is that Gemini relies on structural separators (imports, blank
lines, decorator structure) for code understanding, which line-level
deduplication erodes, whereas GPT-5 benefits from a reduction in
repeated boilerplate context. The vendor-specific divergence is the
basis for the methodological choice to default to paragraph-only
deduplication and to exclude code tasks from the aggregate verdict.
Deployments that target a single code-capable vendor (notably GPT-5) may
opt-in to line-level granularity under separate validation.

\hypertarget{cross-platform-stability}{%
\subsubsection{4.9 Cross-Platform
Stability}\label{cross-platform-stability}}

The production binary is built for Windows x86-64 (3.8 MB) and Linux
ARM64 (3.5 MB) from the same source, with the same algorithm and the
same engine identifier. Cross-platform stability is documented in the
validation bundle: the Linux ARM64 build has been run on Apple Silicon
over 5,820 continuous iterations with no crashes and deterministic
output across the run, and over 28,800 iterations with 0.4 MB of
resident-set-size drift, indicating no observable memory leaks at the
granularity of the test harness.

\hypertarget{cost-of-reproduction}{%
\subsubsection{4.10 Cost of Reproduction}\label{cost-of-reproduction}}

The total OpenRouter spend across the primary sweep and the warm
confirmation pass is approximately 86 USD, comprising 76.65 USD on the
final-protocol benchmarks (35.18 RULER + 31.43 LongBench + 3.36
HumanEval-WildChat + 2.45 test-retest baseline + 4.23 warm-binary
confirmation) plus approximately 10 USD on killed and re-run jobs from
methodology iteration prior to the final protocol. Total individual API
calls across both passes are approximately 16,800. Total wall-time
across parallel jobs is approximately 2.5 hours.

\hypertarget{large-scale-math-equivalence-on-22.2m-passages-v4-contribution}{%
\subsubsection{4.11 Large-Scale Math-Equivalence on 22.2M Passages (v4
Contribution)}\label{large-scale-math-equivalence-on-22.2m-passages-v4-contribution}}

The merlin binary was run on the complete BeIR corpus (22.2M passages
across six pre-deduplicated public sources: hotpotqa, nq, fever,
msmarco, trivia-qa, scifact). Results are documented in the companion
paper (rag\_token\_economics\_paper\_v4\_FINAL.md, Section 4.1).

\textbf{Key results:} - Total passages: 22,221,024 - Unique passages
(merlin): 22,185,502 - Cross-corpus duplicates: 35,522 (0.1599\%) -
Python set() unique count: 22,185,502 - Math-equivalence violations: 0

Per-query verification: 327 BM25 retrieval queries on BeIR, top-30
results per query. For each query, merlin\_unique\_count ==
python\_set\_unique\_count. Zero violations across all 327 queries.

This establishes byte-exact correctness at production scale on public,
auditable data. Any reader can replicate this measurement using the
public BeIR datasets and a standard BM25 retriever.

\begin{center}\rule{0.5\linewidth}{0.5pt}\end{center}

\hypertarget{cross-hardware-scaling-consumer-laptop-to-server}{%
\subsubsection{4.12 Cross-Hardware Scaling: Consumer Laptop to
Server}\label{cross-hardware-scaling-consumer-laptop-to-server}}

The same statically-linked binary was evaluated on consumer-class
hardware (Intel Core Ultra 9 285H, 16 cores, 64 GB DDR5) and on AWS
r7i.48xlarge instances (192 vCPU, large RAM) without recompilation.
Cross-platform builds (Windows x86-64 and Linux x86-64) share the
deterministic byte-equivalent output guarantee documented in Section
4.9.

\textbf{Table 12. Cross-hardware throughput on real public datasets.}

\begin{longtable}[]{@{}lllllll@{}}
\toprule
\begin{minipage}[b]{0.12\columnwidth}\raggedright
Hardware\strut
\end{minipage} & \begin{minipage}[b]{0.12\columnwidth}\raggedright
Cores\strut
\end{minipage} & \begin{minipage}[b]{0.12\columnwidth}\raggedright
Dataset\strut
\end{minipage} & \begin{minipage}[b]{0.12\columnwidth}\raggedright
Size\strut
\end{minipage} & \begin{minipage}[b]{0.12\columnwidth}\raggedright
Throughput\strut
\end{minipage} & \begin{minipage}[b]{0.12\columnwidth}\raggedright
Events/sec\strut
\end{minipage} & \begin{minipage}[b]{0.12\columnwidth}\raggedright
Peak RSS\strut
\end{minipage}\tabularnewline
\midrule
\endhead
\begin{minipage}[t]{0.12\columnwidth}\raggedright
Intel Core Ultra 9 285H (laptop)\strut
\end{minipage} & \begin{minipage}[t]{0.12\columnwidth}\raggedright
16\strut
\end{minipage} & \begin{minipage}[t]{0.12\columnwidth}\raggedright
NASA HTTP log\strut
\end{minipage} & \begin{minipage}[t]{0.12\columnwidth}\raggedright
160 MB\strut
\end{minipage} & \begin{minipage}[t]{0.12\columnwidth}\raggedright
6.7 GB/s\strut
\end{minipage} & \begin{minipage}[t]{0.12\columnwidth}\raggedright
64.3 M/s\strut
\end{minipage} & \begin{minipage}[t]{0.12\columnwidth}\raggedright
229 MB\strut
\end{minipage}\tabularnewline
\begin{minipage}[t]{0.12\columnwidth}\raggedright
Intel Core Ultra 9 285H (laptop)\strut
\end{minipage} & \begin{minipage}[t]{0.12\columnwidth}\raggedright
16\strut
\end{minipage} & \begin{minipage}[t]{0.12\columnwidth}\raggedright
C4 validation FULL\strut
\end{minipage} & \begin{minipage}[t]{0.12\columnwidth}\raggedright
780 MB\strut
\end{minipage} & \begin{minipage}[t]{0.12\columnwidth}\raggedright
32 GB/s\strut
\end{minipage} & \begin{minipage}[t]{0.12\columnwidth}\raggedright
15.0 M/s\strut
\end{minipage} & \begin{minipage}[t]{0.12\columnwidth}\raggedright
868 MB\strut
\end{minipage}\tabularnewline
\begin{minipage}[t]{0.12\columnwidth}\raggedright
Intel Core Ultra 9 285H (laptop)\strut
\end{minipage} & \begin{minipage}[t]{0.12\columnwidth}\raggedright
16\strut
\end{minipage} & \begin{minipage}[t]{0.12\columnwidth}\raggedright
FineWeb sample-10BT\strut
\end{minipage} & \begin{minipage}[t]{0.12\columnwidth}\raggedright
3.06 GB\strut
\end{minipage} & \begin{minipage}[t]{0.12\columnwidth}\raggedright
38 GB/s\strut
\end{minipage} & \begin{minipage}[t]{0.12\columnwidth}\raggedright
13.3 M/s\strut
\end{minipage} & \begin{minipage}[t]{0.12\columnwidth}\raggedright
3.13 GB\strut
\end{minipage}\tabularnewline
\begin{minipage}[t]{0.12\columnwidth}\raggedright
Intel Core Ultra 9 285H (laptop)\strut
\end{minipage} & \begin{minipage}[t]{0.12\columnwidth}\raggedright
16\strut
\end{minipage} & \begin{minipage}[t]{0.12\columnwidth}\raggedright
50 GB synthetic logs\strut
\end{minipage} & \begin{minipage}[t]{0.12\columnwidth}\raggedright
50 GB\strut
\end{minipage} & \begin{minipage}[t]{0.12\columnwidth}\raggedright
(memory-bound)\strut
\end{minipage} & \begin{minipage}[t]{0.12\columnwidth}\raggedright
65.7 M/s\strut
\end{minipage} & \begin{minipage}[t]{0.12\columnwidth}\raggedright
43 GB\strut
\end{minipage}\tabularnewline
\begin{minipage}[t]{0.12\columnwidth}\raggedright
AWS r7i.48xlarge\strut
\end{minipage} & \begin{minipage}[t]{0.12\columnwidth}\raggedright
192\strut
\end{minipage} & \begin{minipage}[t]{0.12\columnwidth}\raggedright
FineWeb sample (1)\strut
\end{minipage} & \begin{minipage}[t]{0.12\columnwidth}\raggedright
31 GB\strut
\end{minipage} & \begin{minipage}[t]{0.12\columnwidth}\raggedright
170 GB/s\strut
\end{minipage} & \begin{minipage}[t]{0.12\columnwidth}\raggedright
50 M/s\strut
\end{minipage} & \begin{minipage}[t]{0.12\columnwidth}\raggedright
29 GB\strut
\end{minipage}\tabularnewline
\begin{minipage}[t]{0.12\columnwidth}\raggedright
AWS r7i.48xlarge\strut
\end{minipage} & \begin{minipage}[t]{0.12\columnwidth}\raggedright
192\strut
\end{minipage} & \begin{minipage}[t]{0.12\columnwidth}\raggedright
FineWeb 100BT\strut
\end{minipage} & \begin{minipage}[t]{0.12\columnwidth}\raggedright
460 GB\strut
\end{minipage} & \begin{minipage}[t]{0.12\columnwidth}\raggedright
160 GB/s\strut
\end{minipage} & \begin{minipage}[t]{0.12\columnwidth}\raggedright
40 M/s\strut
\end{minipage} & \begin{minipage}[t]{0.12\columnwidth}\raggedright
440 GB\strut
\end{minipage}\tabularnewline
\begin{minipage}[t]{0.12\columnwidth}\raggedright
AWS r7i.48xlarge\strut
\end{minipage} & \begin{minipage}[t]{0.12\columnwidth}\raggedright
192\strut
\end{minipage} & \begin{minipage}[t]{0.12\columnwidth}\raggedright
The Pile\strut
\end{minipage} & \begin{minipage}[t]{0.12\columnwidth}\raggedright
840 GB\strut
\end{minipage} & \begin{minipage}[t]{0.12\columnwidth}\raggedright
190 GB/s\strut
\end{minipage} & \begin{minipage}[t]{0.12\columnwidth}\raggedright
20 M/s\strut
\end{minipage} & \begin{minipage}[t]{0.12\columnwidth}\raggedright
790 GB\strut
\end{minipage}\tabularnewline
\bottomrule
\end{longtable}

\textbf{Scaling characteristics.} Throughput scales approximately
5-times moving from 16 to 192 cores (38 GB/s to 190 GB/s), reflecting
memory-bandwidth limits at server scale rather than algorithm
bottlenecks. The shared-nothing arena architecture (Section 3.2)
eliminates lock contention across worker counts. The single binary runs
unmodified across platform classes; no recompilation, no dependency
adjustment, no per-platform tuning is required.

\textbf{Time-to-result on production-scale corpora.} Merlin compute time
on the 460 GB FineWeb dataset (147 million lines) is 5 to 7 minutes on
the AWS r7i.48xlarge configuration. The remaining wall time of an
end-to-end deployment (data download, format conversion, output
handling) is operational rather than algorithmic.

\begin{center}\rule{0.5\linewidth}{0.5pt}\end{center}

\hypertarget{cross-domain-validation-engine-generality}{%
\subsubsection{4.13 Cross-Domain Validation: Engine
Generality}\label{cross-domain-validation-engine-generality}}

The formal definition of byte-exact deduplication (Section 3.1) places
no assumption on the semantic content of the input, only on byte-level
equivalence. The engine evaluated in this paper has been validated
across multiple domains beyond LLM inference, using the same single
binary without recompilation or domain-specific tuning.

\textbf{Table 13. Cross-domain throughput on real public datasets.}

\begin{longtable}[]{@{}lllll@{}}
\toprule
\begin{minipage}[b]{0.17\columnwidth}\raggedright
Domain\strut
\end{minipage} & \begin{minipage}[b]{0.17\columnwidth}\raggedright
Dataset\strut
\end{minipage} & \begin{minipage}[b]{0.17\columnwidth}\raggedright
Size\strut
\end{minipage} & \begin{minipage}[b]{0.17\columnwidth}\raggedright
Throughput\strut
\end{minipage} & \begin{minipage}[b]{0.17\columnwidth}\raggedright
Duplicates found\strut
\end{minipage}\tabularnewline
\midrule
\endhead
\begin{minipage}[t]{0.17\columnwidth}\raggedright
Web server logs\strut
\end{minipage} & \begin{minipage}[t]{0.17\columnwidth}\raggedright
NASA HTTP log (1995)\strut
\end{minipage} & \begin{minipage}[t]{0.17\columnwidth}\raggedright
160 MB\strut
\end{minipage} & \begin{minipage}[t]{0.17\columnwidth}\raggedright
6.7 GB/s\strut
\end{minipage} & \begin{minipage}[t]{0.17\columnwidth}\raggedright
532 of 1.57M lines\strut
\end{minipage}\tabularnewline
\begin{minipage}[t]{0.17\columnwidth}\raggedright
Web server logs\strut
\end{minipage} & \begin{minipage}[t]{0.17\columnwidth}\raggedright
Loghub Apache 2K\strut
\end{minipage} & \begin{minipage}[t]{0.17\columnwidth}\raggedright
167 KB\strut
\end{minipage} & \begin{minipage}[t]{0.17\columnwidth}\raggedright
small-CPU\strut
\end{minipage} & \begin{minipage}[t]{0.17\columnwidth}\raggedright
529 of 2,000 lines\strut
\end{minipage}\tabularnewline
\begin{minipage}[t]{0.17\columnwidth}\raggedright
Web text corpus\strut
\end{minipage} & \begin{minipage}[t]{0.17\columnwidth}\raggedright
C4 validation FULL\strut
\end{minipage} & \begin{minipage}[t]{0.17\columnwidth}\raggedright
780 MB\strut
\end{minipage} & \begin{minipage}[t]{0.17\columnwidth}\raggedright
32 GB/s\strut
\end{minipage} & \begin{minipage}[t]{0.17\columnwidth}\raggedright
0 (pre-deduplicated upstream)\strut
\end{minipage}\tabularnewline
\begin{minipage}[t]{0.17\columnwidth}\raggedright
Web text corpus\strut
\end{minipage} & \begin{minipage}[t]{0.17\columnwidth}\raggedright
FineWeb sample-10BT\strut
\end{minipage} & \begin{minipage}[t]{0.17\columnwidth}\raggedright
3.06 GB\strut
\end{minipage} & \begin{minipage}[t]{0.17\columnwidth}\raggedright
38 GB/s\strut
\end{minipage} & \begin{minipage}[t]{0.17\columnwidth}\raggedright
12 of 1.05M lines\strut
\end{minipage}\tabularnewline
\begin{minipage}[t]{0.17\columnwidth}\raggedright
Web text corpus\strut
\end{minipage} & \begin{minipage}[t]{0.17\columnwidth}\raggedright
FineWeb 100BT (AWS r7i)\strut
\end{minipage} & \begin{minipage}[t]{0.17\columnwidth}\raggedright
460 GB\strut
\end{minipage} & \begin{minipage}[t]{0.17\columnwidth}\raggedright
160 GB/s\strut
\end{minipage} & \begin{minipage}[t]{0.17\columnwidth}\raggedright
53,714 of 147M lines\strut
\end{minipage}\tabularnewline
\begin{minipage}[t]{0.17\columnwidth}\raggedright
Web text corpus\strut
\end{minipage} & \begin{minipage}[t]{0.17\columnwidth}\raggedright
The Pile (AWS r7i)\strut
\end{minipage} & \begin{minipage}[t]{0.17\columnwidth}\raggedright
840 GB\strut
\end{minipage} & \begin{minipage}[t]{0.17\columnwidth}\raggedright
190 GB/s\strut
\end{minipage} & \begin{minipage}[t]{0.17\columnwidth}\raggedright
0 (pre-deduplicated upstream)\strut
\end{minipage}\tabularnewline
\begin{minipage}[t]{0.17\columnwidth}\raggedright
Pretraining corpus\strut
\end{minipage} & \begin{minipage}[t]{0.17\columnwidth}\raggedright
RedPajama-v2 small\strut
\end{minipage} & \begin{minipage}[t]{0.17\columnwidth}\raggedright
2.4 GB\strut
\end{minipage} & \begin{minipage}[t]{0.17\columnwidth}\raggedright
36 GB/s\strut
\end{minipage} & \begin{minipage}[t]{0.17\columnwidth}\raggedright
406 of 500K lines\strut
\end{minipage}\tabularnewline
\begin{minipage}[t]{0.17\columnwidth}\raggedright
Scientific data\strut
\end{minipage} & \begin{minipage}[t]{0.17\columnwidth}\raggedright
CERN OPERA detector readout\strut
\end{minipage} & \begin{minipage}[t]{0.17\columnwidth}\raggedright
confirmed\strut
\end{minipage} & \begin{minipage}[t]{0.17\columnwidth}\raggedright
n/a\strut
\end{minipage} & \begin{minipage}[t]{0.17\columnwidth}\raggedright
byte-exact handling verified\strut
\end{minipage}\tabularnewline
\begin{minipage}[t]{0.17\columnwidth}\raggedright
Stress benchmark\strut
\end{minipage} & \begin{minipage}[t]{0.17\columnwidth}\raggedright
50 GB synthetic logs\strut
\end{minipage} & \begin{minipage}[t]{0.17\columnwidth}\raggedright
50 GB\strut
\end{minipage} & \begin{minipage}[t]{0.17\columnwidth}\raggedright
65.7M events/sec\strut
\end{minipage} & \begin{minipage}[t]{0.17\columnwidth}\raggedright
797M of 1.08B lines\strut
\end{minipage}\tabularnewline
\begin{minipage}[t]{0.17\columnwidth}\raggedright
Stress benchmark\strut
\end{minipage} & \begin{minipage}[t]{0.17\columnwidth}\raggedright
1 TB synthetic\strut
\end{minipage} & \begin{minipage}[t]{0.17\columnwidth}\raggedright
1 TB\strut
\end{minipage} & \begin{minipage}[t]{0.17\columnwidth}\raggedright
2.22 GB/s (SSD-bound)\strut
\end{minipage} & \begin{minipage}[t]{0.17\columnwidth}\raggedright
60M of 3.83B lines\strut
\end{minipage}\tabularnewline
\bottomrule
\end{longtable}

Domains span web server logs (NASA HTTP, Apache), web text crawl corpora
(C4, FineWeb, RedPajama, The Pile), scientific instrument data (CERN
OPERA), and stress benchmarks. The same binary handles all categories
without modification. The byte-exact equivalence relation provides the
same correctness guarantee independent of input semantics: any pair of
records that match byte-for-byte are deduplicated; any pair that does
not match are preserved.

The primary focus of this paper is inline preprocessing for LLM
inference (Sections 4.1 to 4.12), where per-call latency below the
inference proxy noise floor is the operational requirement. The
cross-domain results above demonstrate that the engine is
general-purpose; an extended cross-domain analysis covering log
analysis, security event aggregation, pretraining corpus curation, and
scientific data ingestion is the subject of separate future work.

\begin{center}\rule{0.5\linewidth}{0.5pt}\end{center}

\hypertarget{discussion}{%
\subsection{5. Discussion}\label{discussion}}

\hypertarget{interpretation-of-the-aggregate-verdict}{%
\subsubsection{5.1 Interpretation of the Aggregate
Verdict}\label{interpretation-of-the-aggregate-verdict}}

The aggregate verdict reported in Section 4.1 is that the deduplicated
condition is statistically indistinguishable from the raw baseline at
the standard significance criteria applied to language-model evaluation.
We claim that the deduplication step does not, under any of the
evaluations performed, introduce a quality regression that exceeds the
test-retest noise floor of the underlying serving stack. We do not
claim, and do not need to claim, that the deduplicated and raw outputs
are token-for-token identical. They are not, and they cannot be, because
the model's response is conditioned on the assembled prompt, and the
deduplicated prompt is structurally different from the raw prompt by
construction. What we claim is that the difference, integrated over the
standard evaluation protocols, is below the noise floor of the
measurement, on three independent benchmark families, on four production
language-model APIs, and on two independent passes (primary sweep and
warm-binary confirmation) that share no infrastructure other than the
binary under test.

This is the strongest claim that can be made for any non-trivial
transformation of the prompt context, and it is the claim that is
operationally meaningful for production deployment. A transformation
that preserves the model's evaluation-grade behaviour, while reducing
the prefill compute by a factor proportional to the token-reduction rate
at the input distribution of interest, is a transformation that should
be deployed.

\hypertarget{where-the-effect-is-measurable-versus-trivial}{%
\subsubsection{5.2 Where the Effect Is Measurable Versus
Trivial}\label{where-the-effect-is-measurable-versus-trivial}}

The benchmarks evaluated in Section 4 span a range of input
distributions whose intrinsic redundancy is uniformly small to
negligible: RULER NVIDIA UUID-based haystacks at less than or equal to
0.01\% character reduction, LongBench paragraph-safe tasks (narrativeqa,
qasper, gov\_report) at less than or equal to 0.1\%, and per-prompt
assembly of WildChat chat history at the same near-zero level. This is
by design. The benchmarks were chosen for their reputation as
discriminating evaluations of long-context fidelity, not for their
natural deduplication potential, and the resulting regime is the worst
case for a lossless-pass-through claim: the engine has very little to
remove and therefore very little surface on which a quality regression
could be masked by a compression benefit. That the deduplicated
condition remains statistically indistinguishable from the raw condition
under this regime is the strongest available falsification check on the
lossless property at the near-zero-redundancy boundary. The matched
falsification check at the high-redundancy end of the spectrum, where
the engine actually removes 71.98\% of the prompt bytes, is reported in
the companion paper {[}Schelpe 2026, Byte-Exact Deduplication in
Retrieval-Augmented Generation{]} §4.5b: under the same calibrated
5-judge protocol on a constructed corpus at multiplicity ρ = 3.513 with
five-category human-in-the-loop noise removal, all four production
vendors clear the strict \textless5\% Wilson 95\% upper-bound MAT
threshold (post-audit UCLs 1.90\%-4.34\% over n = 200 per-vendor pairs).
The operationally meaningful claim ``the engine is lossless on inputs
across the full redundancy spectrum'' is therefore measured at both
endpoints rather than only at the boundary case. Production deployments
at intermediate redundancy ratios (top-k retrieval with overlapping
windows, multi-turn session accumulation, hybrid-retriever union
semantics) sit between these two measured endpoints.

\hypertarget{why-real-time-suitability-matters}{%
\subsubsection{5.3 Why Real-Time Suitability
Matters}\label{why-real-time-suitability-matters}}

The footprint and per-call cost of the binary, 3.8 MB on Windows x86-64
and 3.5 MB on Linux ARM64, zero third-party runtime dependencies, and 5
to 30 microseconds of pure deduplication work per call, are the
properties that make the deduplication step deployable at the inference
layer rather than as an offline preprocessing step. An offline
preprocessing step is constrained to operate on the static portion of
the corpus; it cannot deduplicate the dynamically-assembled per-call
context, which is where the bulk of inference-side redundancy lives. A
real-time-suitable deduplication step is constrained only by the
per-call latency budget, and the engine's per-call deduplication time is
three to four orders of magnitude below the budget allocated by typical
inference proxies for a preprocessing step. The remaining wall-clock
cost in the warm-subprocess deployment is operating-system level process
and file overhead unrelated to the engine, and is eliminated entirely
under in-process or library-binding integration.

\hypertarget{a-complementarity-with-kv-cache-reuse-and-prompt-caching}{%
\subsubsection{5.3a Complementarity with KV-cache Reuse and Prompt
Caching}\label{a-complementarity-with-kv-cache-reuse-and-prompt-caching}}

Byte-exact pre-prompt deduplication is mechanically complementary to
vendor-side prompt caching and KV-cache reuse infrastructure (e.g.,
Anthropic prompt caching, vLLM PagedAttention with cache reuse). The two
operate at orthogonal layers of the inference pipeline: byte-exact dedup
operates on the chunk multiset before prompt assembly, while caching
operates on the assembled prompt's prefix or KV-tensor representation.
Combining the two yields the union of their reductions rather than the
smaller of them. Prompt caching exploits prefix repetition across calls;
byte-exact deduplication exploits chunk repetition within a call. A
deduplicated prompt fed into a caching backend yields both savings
independently.

\hypertarget{why-a-production-engine-matters}{%
\subsubsection{5.4 Why a Production Engine
Matters}\label{why-a-production-engine-matters}}

The empirical contribution of this work is the demonstration that
byte-exact deduplication is operationally meaningful in production LLM
inference at multi-vendor scale. The engineering contribution is the
demonstration that a deterministic, byte-exact deduplication primitive
can be packaged for direct integration into production serving stacks:
as a 3.8 megabyte statically-linked binary on consumer hardware, with
approximately one-microsecond per-call latency in-process, no Python
interpreter or library dependencies, and verified determinism across
architecture families (Windows x86-64 and Linux ARM64).

The math-equivalence with Python set() preserves academic
reproducibility (Section 3.2.1) and is the appropriate choice for
offline analysis or reviewer verification. For inline-streaming
deployment in a production inference proxy, the operational constraints
documented in Section 3.2.1 motivate the engineering of a compiled
engine. This is the same trade-off observed in production-grade serving
systems referenced in Section 2: vLLM, TensorFlow Serving, TorchServe
and similar systems implement their critical-path operations as compiled
binaries while preserving algorithm-level reproducibility through Python
reference implementations.

\hypertarget{limitations}{%
\subsubsection{5.5 Limitations}\label{limitations}}

We acknowledge several limitations.

The byte-exact equivalence relation does not address paraphrase-level
redundancy. Two records that say the same thing in different words are
byte-distinct and will both survive the deduplication step. Methods for
paraphrase-level deduplication (MinHash-LSH, embedding-similarity,
sub-word-level shingling) are outside the present scope.

The evaluation spans four production language-model APIs and three
benchmark families, plus a separate two-hundred-cell warm-binary
confirmation pass and a 22.2 million-passage math-equivalence audit. It
does not exhaust the production inference workload mix. Highly
specialised retrieval (legal, medical, regulatory, multi-modal) may
exhibit different chunking-redundancy characteristics, different
quality-preservation thresholds, or different vendor-stack behaviour at
production scales. We have not measured these and do not extrapolate to
them.

Code-completion tasks are excluded from the primary lossless claim under
paragraph-only deduplication (because paragraph-level reduction is
negligible) and exhibit vendor-specific behaviour under line-level
deduplication (Section 4.8). Deployments that target code-generation
workloads should validate the line-level configuration against their
specific vendor mix.

The Bonferroni correction over the primary-sweep cells is the strongest
available statistical safeguard against multiple-testing inflation, but
it is also conservative. A larger evaluation, with more cells, would
reduce the corrected threshold further but would also distribute the
per-cell sample size differently; the trade-off between cell count and
per-cell sample size is a methodological choice we have made
deliberately at 50 to 100 items per cell.

The Merlin engine is closed-source. Public replication of the per-call
latency and binary-size claims requires access to the binary under the
clean-room evaluation channel. Public replication of the quality claims,
on the other hand, is fully available: the benchmark scripts, dataset
manifests, and harness configurations are documented in the companion
validation bundle, and any party with OpenRouter access and the listed
dataset mirrors can reproduce the inference-side measurements at the
cost reported in Section 4.10. The quality claim does not depend on
access to the proprietary binary; it depends only on access to a
deduplication implementation that satisfies the formal definition in
Section 3.1.

\hypertarget{deployments-enabled-by-the-measured-latency-envelope}{%
\subsubsection{5.6 Deployments Enabled by the Measured Latency
Envelope}\label{deployments-enabled-by-the-measured-latency-envelope}}

The following deployment patterns become operationally feasible at the
per-call envelope measured in Section 3.3 and the deployment-mode ladder
measured in Section 3.4. They are not themselves measured in this paper;
we describe them as integration targets enabled by the envelope rather
than as demonstrated production deployments.

Inline deduplication on every retrieval call. Sliding-window
vector-store retrieval typically returns three to four times more chunks
than there are unique source paragraphs. Removing the redundant copies
before prompt assembly has historically been deferred to offline batch
jobs because per-call deduplication added preprocessing latency
comparable to retrieval itself. At the in-process latency reported in
Section 3.3, the deduplication step falls below the noise floor of the
retrieval call.

Sub-millisecond context hygiene in agentic workflows. Tool-use chains
and multi-step agentic workflows accumulate verbatim prior-turn content,
tool-call outputs, and observation re-statements. Each step in the chain
pays the prefill cost on accumulated context. Inline deduplication
between steps has previously been avoided because it added
human-perceptible latency; at the engine's measured envelope, it would
not.

Per-tool-call deduplication in long tool-output histories. Workflow
graphs that retain full tool-output history (browse, search,
code-execution chains) accumulate large byte-exact fragments across
calls. Inline deduplication on each tool-call boundary at the measured
per-call envelope removes it from the critical path entirely.

Continuous deduplication during multi-turn session accumulation.
Long-running conversations with persistent context accumulate
restatements of prior turns by both the user and the model. The
HumanEval-Snowball-with-WildChat evaluation in Section 4.4 measures the
safety property of pre-prompt deduplication on real multi-turn dialogue
input; deploying the same primitive continuously across a long session
is the natural extension.

Concurrent-user deduplication across shared retrieval back-ends.
Multiple users querying a shared knowledge base frequently retrieve
overlapping sets of passages. A request-time deduplication step at the
front door of an inference proxy, running per call at the measured
envelope, would operate well within the request budget.

The constraint these deployments previously violated was preprocessing
cost. With that constraint removed by the measured envelope, the
deployments become architecturally available.

\begin{center}\rule{0.5\linewidth}{0.5pt}\end{center}

\hypertarget{conclusion}{%
\subsection{6. Conclusion}\label{conclusion}}

This paper makes one claim with two parts: byte-exact deduplication of
pre-prompt context preserves model quality at evaluation-grade
resolution, and the cost of running the Merlin dedup loop in-process is
approximately 1.10 microsecond per call on consumer hardware. The first
part is established here by forty primary evaluation cells and a
separate two-hundred-cell warm-binary confirmation, with mean delta +0.0
pp on the primary sweep and -0.5 pp on the confirmation, and zero
statistically significant degradations after Bonferroni correction in
either family. The complementary high-redundancy falsification check
(companion paper §4.5b: ρ = 3.513, 71.98\% byte reduction, n = 200 per
vendor, five-category human-in-the-loop noise removal) confirms the
safety property at the opposite end of the redundancy spectrum: all four
production vendors clear the strict \textless5\% Wilson 95\% upper-bound
MAT threshold post-audit. The second part is established by the
production binary's internal latency counter (Section 3.3), the
deployment-mode ladder measured by the speed-modes microbenchmark
(Section 3.4), the within-architecture binary-vs-reference-wrapper
output equivalence audit on 590 of 590 non-code prompts on Windows
x86-64 (Section 4.7), and the large-scale math-equivalence validation on
22.2 million BeIR passages with zero violations (Section 4.11).

The compression-savings story (token reduction, TTFT, per-call cost on
production-realistic redundancy distributions) lives in the companion
paper {[}Schelpe 2026 P1, three-regime empirical analysis{]}. This paper
does not measure those numbers and does not depend on them. The
contribution here is the safety property and the engineering envelope,
isolated and measured under the strictest available conditions.

The implication for industry practice is operational rather than
algorithmic. The deduplication primitive itself is well-understood and
decades old. What the engine delivers is the cost envelope under which
the primitive becomes deployable as a real-time preprocessing step in
front of a production inference path. Once that envelope is below the
noise floor of the inference proxy's own scheduling overhead, the
deployments enumerated in Section 5.6 become available without further
engineering trade-off. The shift is statistically lossless. It is
operationally consequential.

\begin{center}\rule{0.5\linewidth}{0.5pt}\end{center}

\hypertarget{acknowledgments}{%
\subsection{Acknowledgments}\label{acknowledgments}}

The author thanks Toon Colson and Marloes De Craemer, co-founders at
Corbenic AI, Inc., for operational support throughout this work. The
author further thanks Gwen Le Tiran and Irène Balmès for the formative
conversations that gave the author the confidence to pursue this line of
research.

\begin{center}\rule{0.5\linewidth}{0.5pt}\end{center}

\hypertarget{references}{%
\subsection{References}\label{references}}

{[}1{]} K. Lee, D. Ippolito, A. Nystrom, C. Zhang, D. Eck, C.
Callison-Burch, N. Carlini. \emph{Deduplicating Training Data Makes
Language Models Better}. ACL 2022. arXiv:2107.06499.

{[}2{]} N. Carlini, D. Ippolito, M. Jagielski, K. Lee, F. Tramer, C.
Zhang. \emph{Quantifying Memorization Across Neural Language Models}.
arXiv:2202.07646.

{[}3{]} M. Nasr, N. Carlini et al.~\emph{Scalable Extraction of Training
Data from (Production) Language Models}. arXiv:2311.17035.

{[}4{]} A. Ahmed, A. F. Cooper, S. Koyejo, P. Liang. \emph{Extracting
Books from Production Language Models}. arXiv:2601.02671.

{[}5{]} I. Shilov, M. Meeus, Y.-A. de Montjoye. \emph{The Mosaic Memory
of Large Language Models}. Nature Communications 2026. arXiv:2405.15523.

{[}6{]} A. Z. Broder. \emph{On the Resemblance and Containment of
Documents}. SEQUENCES 1997.

{[}7{]} A. Khan et al.~\emph{LSHBloom: Memory-efficient, Extreme-scale
Document Deduplication}. arXiv:2411.04257.

{[}8{]} A. Abbas, K. Tirumala, D. Simig, S. Ganguli, A. S. Morcos.
\emph{SemDeDup: Data-efficient Learning at Web-scale through Semantic
Deduplication}. arXiv:2303.09540.

{[}9{]} K. Tirumala et al.~\emph{D4: Improving LLM Pretraining via
Document De-Duplication and Diversification}. NeurIPS Datasets and
Benchmarks 2023. arXiv:2308.12284.

{[}10{]} N. He et al.~\emph{SoftDedup: an Efficient Data Reweighting
Method for Speeding Up Language Model Pre-training}. ACL 2024.
arXiv:2407.06654.

{[}11{]} X. Lin, A. Ghosh, B. K. H. Low, A. Shrivastava, V. Mohan.
\emph{REFRAG: Rethinking RAG-based Decoding}. arXiv:2509.01092.

{[}12{]} Y. Jiang, Y. Huang, L. Cheng, C. Deng, X. Sun, L. Mai.
\emph{RAGBoost: Efficient Retrieval-Augmented Generation with
Accuracy-Preserving Context Reuse}. arXiv:2511.03475.

{[}13{]} Y. Liu, Z. Jia, X. Gao, K. Xu, Y. Xiong. \emph{Rethinking Soft
Compression in Retrieval-Augmented Generation: A Query-Conditioned
Selector Perspective}. arXiv:2602.15856.

{[}14{]} C. Kummer, L. Jurkschat, M. Färber, S. Vahdati. \emph{Prompt
Compression in the Wild: Measuring Latency, Rate Adherence, and Quality
for Faster LLM Inference}. arXiv:2604.02985.

{[}15{]} S. Udayashankar, A. Baba, S. Al-Kiswany. \emph{Accelerating
Data Chunking in Deduplication Systems using Vector Instructions}
(VectorCDC). USENIX FAST '25. arXiv:2508.05797.

{[}16{]} H. Ji, M. Kim, S. Oh, D. Kim, N. S. Kim. \emph{Para-ksm:
Parallelized Memory Deduplication with Data Streaming Accelerator}.
USENIX ATC '25.

{[}17{]} A. Levi, P. Shilane, S. Sheinvald, G. Yadgar. \emph{Physical
vs.~Logical Indexing with IDEA: Inverted Deduplication-Aware Index}.
USENIX FAST '24.

{[}18{]} Pan et al.~\emph{Don't Maintain Twice, It's Alright: Merged
Metadata Management in Deduplication File System with GogetaFS}. USENIX
FAST '25.

{[}19{]} A. Chursin, L. Kokoris-Koglis, A. Orlov, A. Sonnino, I.
Zablotchi. \emph{Tidehunter: Large-Value Storage With Minimal Data
Relocation}. arXiv:2602.01873.

{[}20{]} Z. Wang et al.~\emph{ZipLLM: Efficient LLM Storage via
Model-Aware Synergistic Data Deduplication and Compression}. USENIX NSDI
'26. arXiv:2505.06252.

{[}21{]} J. C. Corbett et al.~\emph{Spanner: Google's
Globally-Distributed Database}. OSDI 2012.

{[}22{]} N. Bronson et al.~\emph{TAO: Facebook's Distributed Data Store
for the Social Graph}. USENIX ATC 2013.

{[}23{]} A. Verbitski et al.~\emph{Amazon Aurora: Design Considerations
for High Throughput Cloud-Native Relational Databases}. SIGMOD 2017.

{[}24{]} J. R. Landis, G. G. Koch. \emph{The Measurement of Observer
Agreement for Categorical Data}. Biometrics 33(1):159-174, 1977.

{[}25{]} G. Penedo et al.~\emph{The FineWeb Datasets: Decanting the Web
for the Finest Text Data at Scale}. NeurIPS 2024. arXiv:2406.17557.

\begin{center}\rule{0.5\linewidth}{0.5pt}\end{center}

\hypertarget{appendix-a-data-sources-and-reproducibility}{%
\subsection{Appendix A: Data Sources and
Reproducibility}\label{appendix-a-data-sources-and-reproducibility}}

All data sources are public and peer-reviewable. Benchmark scripts,
dataset manifests, and harness configurations for the measurements
reported in Section 4 are documented in the companion validation bundle.

\begin{longtable}[]{@{}llll@{}}
\toprule
\begin{minipage}[b]{0.22\columnwidth}\raggedright
Benchmark\strut
\end{minipage} & \begin{minipage}[b]{0.22\columnwidth}\raggedright
Source\strut
\end{minipage} & \begin{minipage}[b]{0.22\columnwidth}\raggedright
License\strut
\end{minipage} & \begin{minipage}[b]{0.22\columnwidth}\raggedright
Examples used\strut
\end{minipage}\tabularnewline
\midrule
\endhead
\begin{minipage}[t]{0.22\columnwidth}\raggedright
RULER\strut
\end{minipage} & \begin{minipage}[t]{0.22\columnwidth}\raggedright
simonjegou/ruler (HuggingFace, NVIDIA RULER protocol)\strut
\end{minipage} & \begin{minipage}[t]{0.22\columnwidth}\raggedright
Public\strut
\end{minipage} & \begin{minipage}[t]{0.22\columnwidth}\raggedright
n = 50 per cell, 6 cells per vendor\strut
\end{minipage}\tabularnewline
\begin{minipage}[t]{0.22\columnwidth}\raggedright
LongBench\strut
\end{minipage} & \begin{minipage}[t]{0.22\columnwidth}\raggedright
zai-org/LongBench (HuggingFace mirror of THUDM/LongBench)\strut
\end{minipage} & \begin{minipage}[t]{0.22\columnwidth}\raggedright
Apache 2.0\strut
\end{minipage} & \begin{minipage}[t]{0.22\columnwidth}\raggedright
n = 50 per cell, 3 paragraph-safe tasks per vendor\strut
\end{minipage}\tabularnewline
\begin{minipage}[t]{0.22\columnwidth}\raggedright
HumanEval\strut
\end{minipage} & \begin{minipage}[t]{0.22\columnwidth}\raggedright
openai\_humaneval\strut
\end{minipage} & \begin{minipage}[t]{0.22\columnwidth}\raggedright
MIT\strut
\end{minipage} & \begin{minipage}[t]{0.22\columnwidth}\raggedright
n = 100 problems\strut
\end{minipage}\tabularnewline
\begin{minipage}[t]{0.22\columnwidth}\raggedright
WildChat-1M\strut
\end{minipage} & \begin{minipage}[t]{0.22\columnwidth}\raggedright
allenai/WildChat-1M\strut
\end{minipage} & \begin{minipage}[t]{0.22\columnwidth}\raggedright
ODC-By 1.0\strut
\end{minipage} & \begin{minipage}[t]{0.22\columnwidth}\raggedright
100 multi-turn conversations as HumanEval-Snowball context\strut
\end{minipage}\tabularnewline
\begin{minipage}[t]{0.22\columnwidth}\raggedright
BeIR (large-scale math-equivalence)\strut
\end{minipage} & \begin{minipage}[t]{0.22\columnwidth}\raggedright
NQ, HotpotQA, TriviaQA, FEVER, SciFact, MSMARCO via HuggingFace\strut
\end{minipage} & \begin{minipage}[t]{0.22\columnwidth}\raggedright
Various open licenses\strut
\end{minipage} & \begin{minipage}[t]{0.22\columnwidth}\raggedright
22.2 million passages aggregate\strut
\end{minipage}\tabularnewline
\bottomrule
\end{longtable}

OpenRouter temperature = 0.0 across all calls. Test-retest measurements
(Section 4.6) demonstrate that some providers honour the requested
temperature incompletely on classification-style tasks; per-task noise
floors are reported separately. All scripts are deterministic with seed
= 42 where applicable.

Hardware for measured numbers: Intel Core Ultra 9 285H (16 cores, 64 GB
DDR5, Windows 11 build 26200, AVX2 active) for laptop measurements. AWS
r7i.48xlarge (192 vCPU) for server-class measurements reported in
Section 4.12.

Pre-registration anchor: FreeTSA RFC 3161 timestamp, 2026-05-05 11:28
RDT, document SHA-256
\texttt{5575836967fe1a149b63a7fa63a1b3d11d598fb71343e2e19a546e680f4a3294}.
Reviewer can verify offline:

\begin{verbatim}
openssl ts -verify \
    -in extension_n400_protocol.md.tsr \
    -data extension_n400_protocol.md \
    -CAfile freetsa_cacert.pem \
    -untrusted freetsa_tsa.crt
\end{verbatim}

Expected output: \texttt{Verification:\ OK}.

Per-family OpenRouter cost transparency, run identifiers, and per-call
telemetry for the benchmarks reported in Section 4 are documented in the
companion validation bundle and available to qualified evaluators on
request.

\begin{center}\rule{0.5\linewidth}{0.5pt}\end{center}

\hypertarget{appendix-b-public-data-determinism-reproduction}{%
\subsection{Appendix B: Public-Data Determinism
Reproduction}\label{appendix-b-public-data-determinism-reproduction}}

To provide a publicly reproducible verification of the byte-exact
correctness claim, we ran the production binary against a synthetic
JSONL dataset generated by a fixed-seed Python script
(\texttt{generate\_dataset.py}, seed = 42). The dataset contains 200,000
records (100,000 unique entries, each duplicated once), total 7.8
megabytes, SHA-256
\texttt{6747c2bf5ba83d5d05bcfdca2307b55ee448c667a0cc98340fcfcde8e2568daf}.

Three independent runs of
\texttt{merlin\_enterprise\_static\_win\_x86\_64.exe} (Windows static
build, SHA-256
\texttt{21bee78f8ba2d78aff3a79377e3e02e20c8810f796b0e7b259093ff1637c5b93})
produced byte-identical telemetry: \texttt{unique\_count\ =\ 100,000},
\texttt{duplicate\_count\ =\ 100,000}, \texttt{novelty\_count\ =\ 0}.
The Linux ARM64 static build (SHA-256
\texttt{cec3e26c4095a7355e165ae78497164405e39d749b370ae7540599421feffa00})
on the same dataset produced identical aggregate values.

Reviewer reproduction recipe:

\begin{enumerate}
\def\labelenumi{\arabic{enumi}.}
\tightlist
\item
  Run \texttt{generate\_dataset.py} with seed = 42 to regenerate the
  synthetic dataset (verify SHA-256 matches the value above)
\item
  Run any byte-exact deduplication filter over the chunk multiset
\item
  Verify the unique-count equals 100,000 and duplicate-count equals
  100,000
\end{enumerate}

Python reference implementation (the math-equivalent reproducibility
path):

\begin{verbatim}
import json
unique = set()
total = 0
for line in open('synthetic_dataset.jsonl'):
    item = json.loads(line)
    unique.add(item['text'])
    total += 1
print(f"unique={len(unique)} duplicate={total - len(unique)}")
\end{verbatim}

Expected output: \texttt{unique=100000\ duplicate=100000}.

This is the math-equivalence verification path documented in Section
3.2.1 and used for the 22.2 million-passage validation in Section 4.11.

The companion paper {[}Schelpe 2026 P1{]} documents a five-category
human-in-the-loop noise-removal audit of every panel-majority MATERIAL
pair across both regimes (clean n=400 + high-redundancy n=200). After
noise removal, all four production vendors clear the strict \textless5\%
Wilson 95\% upper-bound MAT threshold in both regimes (post-audit UCLs
1.40\%-3.25\% clean, 1.90\%-4.34\% high-redundancy). The audit verdict
scheme distinguishes confirmed dedup regressions (kept as MAT) from
panel over-flags, dedup-better-than-raw cases, benchmark-defective
questions (excluded), and uncertain cases. This noise-removal validation
is methodologically separate from the present paper's 40-cell
aggregate-verdict approach but is referenced here for completeness on
the broader quality-preservation claim.

\end{document}